%% file: main.tex
\documentclass[11pt,a4paper]{article}

\usepackage{xpeng-template}

\usepackage{tikz}
\usepackage{pgfplots}
\usetikzlibrary{plotmarks, calc, positioning, fit, arrows.meta, backgrounds}
\pgfplotsset{compat=1.18}
\usepackage{placeins}
\usepackage{float}
\usepackage{flafter}
\usepackage{hyperref}


\definecolor{PolicyPoint}{HTML}{0077BB}
\definecolor{SelectedPolicy}{HTML}{CC3311}
\definecolor{LocS2}{HTML}{4E46C9}      
\definecolor{LocS1}{HTML}{0E7A6E}      
\definecolor{LocS0}{HTML}{B25E12}      
\definecolor{LocHandoff}{HTML}{5E6A82} 
\definecolor{LocInk}{HTML}{1B2230}
\definecolor{LocInkSoft}{HTML}{5A6577}
\definecolor{SimTint}{HTML}{0E63A6}    
\definecolor{RealTint}{HTML}{B25E12}   

\input{main/macros}

\xpengaffil{XPENG Robotics}
\xpengcorrespondence{Jie Chen}{chenj81@xiaopeng.com}

\title{DeepInsight II: One Trace from Benchmark to Robot}
\author{Siyi Li, Yuchen Kang, Wuliang Wang, Zhengjie Zhang, Jiangpin Liu, Jianhao Yao, Jie Chen\textsuperscript{\textdagger}}
\date{}

\begin{document}

\begin{xpenghero}
\input{main/abstract}
\end{xpenghero}

\xpengbodystyle

\input{main/introduction}
\input{main/related_work}
\input{main/method}
\input{main/experiments}
\input{main/conclusion}

\bibliographystyle{unsrtnat}
\begingroup
\fontsize{10pt}{11pt}\selectfont
\setlength{\bibsep}{0.1em plus 0.05em}
\IfFileExists{refs.bib}{\bibliography{refs}}{}
\endgroup


\end{document}

%% file: main/macros.tex
\newcommand{\di}{DeepInsight\xspace}
\newcommand{\ssystem}[1]{System~#1}
\newcommand{\stwo}{System~2\xspace}
\newcommand{\sone}{System~1\xspace}
\newcommand{\szero}{System~0\xspace}

\newcommand{\refv}[1]{{\color{black!55}#1}}

%% file: main/abstract.tex
\begin{abstract}
Across a Physical AI stack, evaluation maturity is inversely aligned with
deployment risk: foundation models enjoy mature, standardized harnesses, while
the embodied layers on which deployment actually turns remain fragmented across
benchmark-specific simulators, embodiments, and interfaces. The first
DeepInsight report (v1) unified evaluation across this stack behind three
abstractions---task, resource, and result---but its quantitative evidence
centered on the foundation-model layer; navigation and manipulation (System 1)
and whole-body control (System 0) remained simulation case studies, and
physical execution was outside its empirical scope. DeepInsight II keeps that
substrate fixed and quantifies the embodied half. First, it reproduces
released-checkpoint references across two navigation and four manipulation
benchmarks under their native protocols. Second, MotionBench places four
released whole-body controllers under one workload and metric contract, then
carries a qualified within-family cohort from parallel simulation to matched
real-robot trials in which simulated and physical rollouts share a parent
trace identity while retaining execution-domain-specific records, making the
sim-to-real gap a native reduction rather than a reconciliation across
toolchains. Third, a composed System 2--1--0 study extends trace localization
into five evidence-grounded handoff labels, each mapped to a concrete repair
action, with a measured repairability criterion and physical episodes testing
the same attribution under hardware-observable state. The contribution is
therefore not a new evaluation architecture, but empirical continuity from
benchmark execution to matched robot evidence and repair-oriented diagnosis.
\end{abstract}

%% file: main/introduction.tex
\section{Introduction}
\label{sec:intro}
\suppressfloats[t]

\begin{figure}[t]
  \centering
  \includegraphics[width=\linewidth]{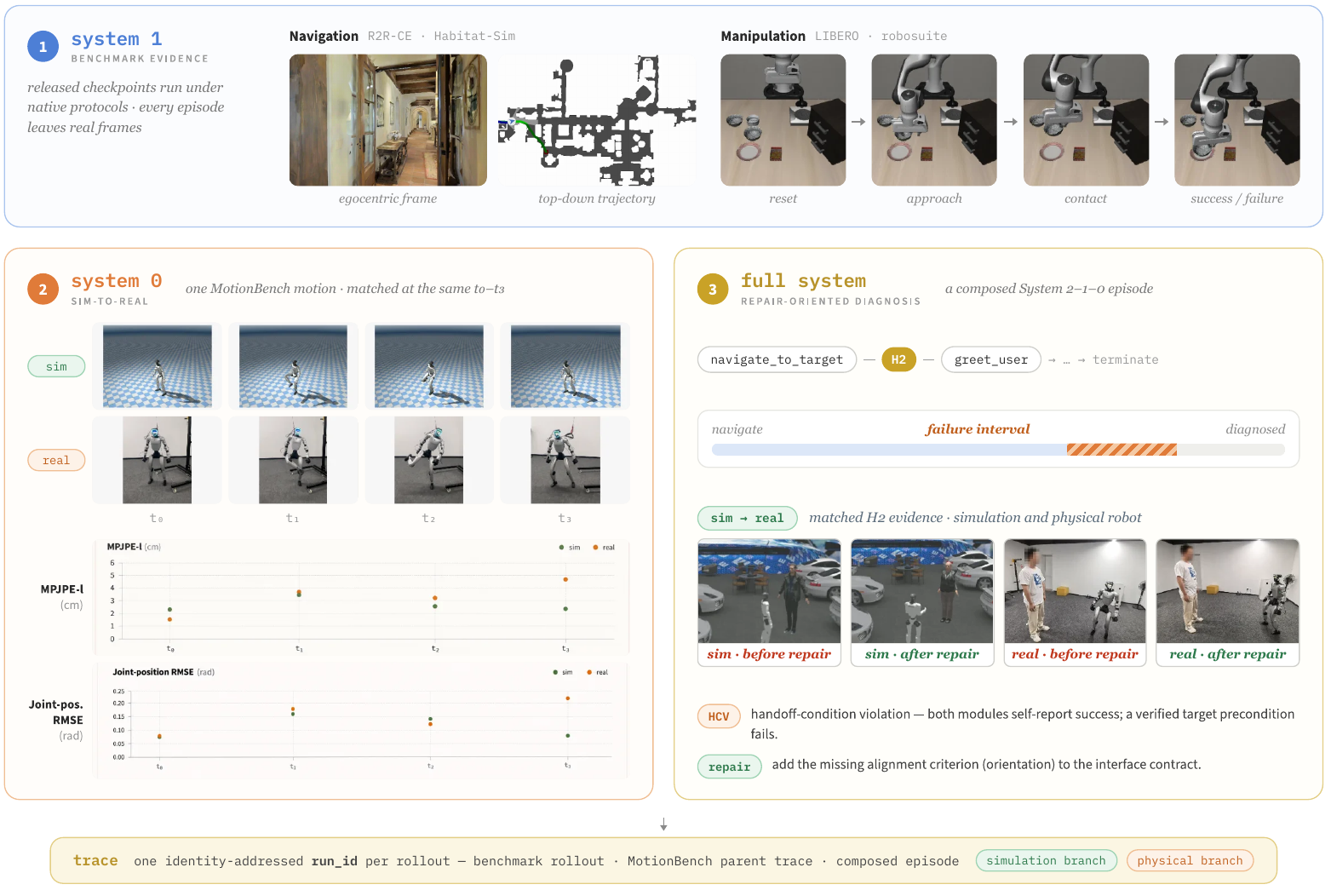}
  \caption{The embodied evidence of this report, on one runtime: \sone
  checkpoints under native benchmark protocols, a \szero motion matched
  between simulation and the physical robot, and a composed
  \ssystem{2--1--0} episode diagnosed down to a concrete repair and re-tested
  in simulation and on the physical robot. Representative H2 frames show the
  state before and after the orientation repair. Every rollout writes into one
  identity-addressed trace, so all three kinds of evidence share a single
  result representation.}
  \label{fig:teaser}
\end{figure}

A policy headed for hardware is evaluated most rigorously where the stakes
are lowest. Its semantic reasoning (\stwo)---the layer farthest from physical
execution, increasingly packaged as a dedicated robot
brain~\citep{li2026athenabrain}---runs through mature harnesses: standardized
datasets, shared inference backends, published protocols. Its navigation and manipulation
skills (\sone) are scored by whichever benchmark each was trained against,
each binding its own simulator, embodiment, and policy interface. And its
whole-body controller (\szero)---the layer whose failures are
physical---is evaluated through per-lab harnesses whose readings rarely
transfer. The layers on which deployment actually turns are the ones
evaluated with the least shared infrastructure. \figref{fig:teaser} previews
the response this report develops.

The difficulty is structural rather than a matter of packaging. Embodied
episodes are closed-loop: every action changes the observation presented at
the next step, so evaluation is a stateful rollout against a live simulator or
robot rather than a batched pass over a static
dataset~\citep{savva2019habitat,liu2023libero,mittal2023isaaclab}. The
resulting heterogeneity has three coupled forms: \emph{workloads} differ in
task and metric semantics; \emph{systems under test} differ in architecture,
checkpoint, controlled DoF, and native input--output interface; and
\emph{execution substrates} range from parallel simulators to serial,
safety-constrained physical robots. Changing any one of these typically
changes the harness and the record against which the result is interpreted:
readings produced by separate harnesses do not automatically share episode
identity, execution semantics, or reduction rules, and aggregate success hides
the local failures and simulation--reality discrepancies that govern
deployment. Recent infrastructure efforts isolate dependencies and federate
benchmark servers~\citep{vlaevalharness}, but typically leave each benchmark's
episode loop and result representation intact; real-robot evaluation is then
joined post hoc, if at all. The objective, therefore, is not to erase this
fragmentation with one universal benchmark or embodiment, but to make the
heterogeneity explicit while preserving comparable evaluation semantics.

The first \di report~\citep{deepinsight} (v1) established the substrate for
this purpose: one \texttt{reset}/\texttt{step} episode driver, one
\texttt{acquire}/\texttt{release} resource protocol, and one
identity-addressed trace across the \ssystem{2--1--0} stack. Its quantitative
evidence, however, centered on \stwo; the embodied layers remained
simulation-only case studies, and physical execution was left outside the
empirical scope. \di II keeps that substrate fixed and closes the empirical
gap. \tabref{tab:v1-v2-delta} states the delta.

\begin{table}[t]
\centering
\scriptsize
\setlength{\tabcolsep}{4pt}
\renewcommand{\arraystretch}{1.25}
\caption{Scope delta from \di v1~\citep{deepinsight} to \di II. The core
substrate is inherited unchanged; \di II is an empirical and
execution-domain extension of v1, not a replacement of its architecture.}
\label{tab:v1-v2-delta}
\begin{tabularx}{\linewidth}{p{0.16\linewidth} X X}
\toprule
\textbf{Dimension}
& \textbf{\di v1 established}
& \textbf{\di II adds} \\
\midrule
Core substrate
& Task, resource, and result: one \texttt{reset}/\texttt{step} driver, one
  \texttt{acquire}/\texttt{release} protocol, one identity-addressed trace.
& \textbf{No new abstraction}; embodied benchmarks and a physical robot bind
  to the same substrate.
\\
\stwo
& Quantitative fidelity, throughput, ablations, and multi-node scaling.
& Inherited; not re-benchmarked here.
\\
\sone
& Simulation-only coverage case studies.
& Native-protocol fidelity: released checkpoints on two navigation and four
  manipulation benchmarks.
\\
\szero
& Trace-backed screening of internal policies, in simulation only.
& MotionBench: four released WBC models under one frozen contract.
\\
Simulation to hardware
& Outside the empirical scope.
& Matched sim and real-robot trials under one parent trace identity, with
  domain-specific records.
\\
Full-system diagnosis
& Task-, subgoal-, and subsystem-level failure localization, in simulation.
& Five handoff labels, each mapped to a repair; measured repairability;
  tested on physical episodes.
\\
\bottomrule
\end{tabularx}
\end{table}

\paragraph{Contributions.}
This paper contributes evaluation infrastructure and empirical evidence, not
new navigation, manipulation, or control policies: native benchmark protocols
remain intact behind adapters, while execution, resource management, trace
identity, and cross-domain reduction are shared. The rest of the paper
follows the three empirical extensions in \tabref{tab:v1-v2-delta}.
\secref{sec:system1} tests native-protocol fidelity across six \sone
benchmarks with released checkpoints. \secref{sec:system0} freezes a common
\szero contract, compares four released WBC models in simulation, and carries
a qualified cohort into matched real-robot trials. \secref{sec:full-system}
extends shared-trace localization into handoff-specific attribution, a
measured repairability criterion separating interface-repairable boundary
errors from action-space limitations, and physical episodes that test the
same attribution under hardware-observable state.

%% file: main/related_work.tex
\section{Related Work}
\label{sec:related}

\paragraph{Embodied benchmarks.}
The benchmarks this paper brings onto one runtime are drawn from three
ecosystems, and we survey them as the workload to be carried rather than as
alternatives to compete with. In navigation, the instruction-following line
(R2R~\citep{anderson2018r2r}, extended to continuous environments by
VLN-CE~\citep{krantz2020vlnce} and to path-fidelity metrics by
RxR~\citep{ku2020rxr}) supplies the workloads and metric suites this paper
carries within the Habitat simulator family. In manipulation, LIBERO's clean
suites~\citep{liu2023libero} are extended by perturbation and generalization
variants such as LIBERO-Plus~\citep{fei2026liberoplus} and
LIBERO-PRO~\citep{liberopro2025}. Other benchmarks cover real-to-sim transfer,
dual-arm control, mobile household tasks, long-horizon evaluation, and
scene-rich environments~\citep{li2024simplerenv,chen2025robotwin2,
robocasa365,mees2022calvin,james2020rlbench,nasiriany2024robocasa}. At
whole-body control,
HumanoidBench~\citep{sferrazza2024humanoidbench} and the Isaac Lab
ecosystem~\citep{mittal2023isaaclab} anchor GPU-parallel locomotion
evaluation. The motion-tracking controllers this paper actually evaluates
descend from the motion-imitation
line~\citep{luo2023phc,tessler2024maskedmimic}: each released
tracker---SONIC~\citep{luo2026sonic},
HoloMotion~\citep{chen2026holomotion}---ships its own evaluation protocol,
with success predicates, tracking-error conventions, and reference-motion
sets that do not interchange across releases; on the training side,
Athena-WBC~\citep{jiang2026athenawbc} argues that long-tail tracking failures
reflect a mismatch between motion demands and learned capability rather than
raw exposure. MotionBench (\secref{sec:system0}) freezes one such contract
across trackers, with capability-grouped reporting aimed at the same axis. \di
preserves each benchmark's protocol---suites, splits,
embodiment, success criteria---and unifies the runtime, resource layer, and
trace beneath them; the fragmentation documented above is, in effect, this
paper's workload specification.

\paragraph{Running embodied benchmarks.}
Most of these benchmarks are run through the native harness each one ships.
The closest infrastructure effort to ours at \sone is the VLA evaluation
harness of \citet{vlaevalharness}, which decouples models from benchmarks by
running each benchmark in its own container against a GPU-resident model
server over a WebSocket message protocol. Batched inference, task sharding,
and result aggregation ride on that separation, dissolving the dependency
conflicts that make embodied benchmarks notoriously non-co-installable. We
adopt the same model-server view of the policy under test in our resource
binding (\secref{sec:method}); beyond such a federation, the shared runtime
adds one episode driver, one result identity across benchmarks, and a real
robot attached behind the same resource protocol---the distinction between
coordination and a shared runtime remains, as in
v1~\citep{deepinsight}, structural rather than nominal. Isaac
Lab~\citep{mittal2023isaaclab} supplies parallel-physics infrastructure that
our \szero workloads build on, and RoboArena~\citep{roboarena2025} makes
distributed real-world evaluation a first-class venue; neither, by design,
joins matched simulated and physical rollouts under one trace identity, which
is the property our sim-to-real study depends on.

\paragraph{Measuring the sim-to-real relation.}
A recent line establishes the predictive validity of simulation for real-robot
evaluation---pairing simulated and physical rollouts, comparing model
orderings and matched errors, and arguing that simulation-based estimators
suffice for checkpoint
selection~\citep{li2024simplerenv,gsworld2025,postconvergence2025,realissim2025,leggedgap2025}.
A complementary construction line narrows the gap rather than measuring it,
rebuilding simulator scenes from sparse real captures around a
robot-executable proxy~\citep{fang2026r2sego}.
We inherit from the measurement line rather than compete with it: the correspondence
analysis of \secref{sec:system0} descends from its metric families, and its
qualification logic shapes our hardware admission gate. What changes is the
domain and the substrate: the line has centered on manipulation policies and
locomotion checkpoints, whereas \secref{sec:system0} asks its question for
whole-body motion-tracking controllers---and matched simulation and hardware
branches share one trace identity and evaluation contract, so the
correspondence evidence is produced by the runtime rather than reconstructed
in a bespoke pipeline beside it.

%% file: main/method.tex
\section{Same Abstractions, New Backends}
\label{sec:method}

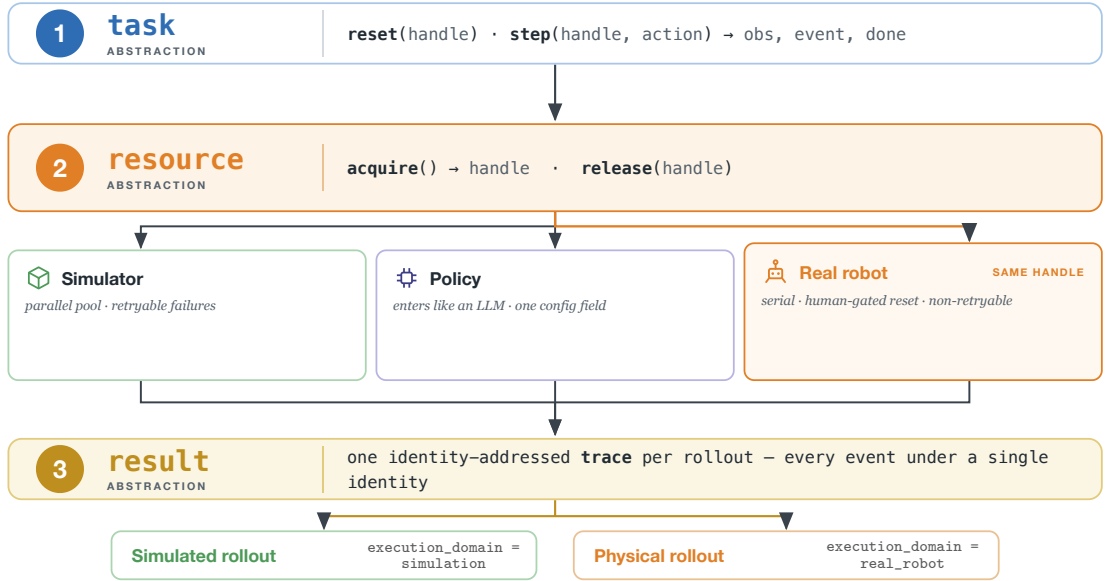
\begin{figure}[t]
  \centering
  \begin{tikzpicture}
    \node[inner sep=0] (overview) {\includegraphics[width=0.92\linewidth]{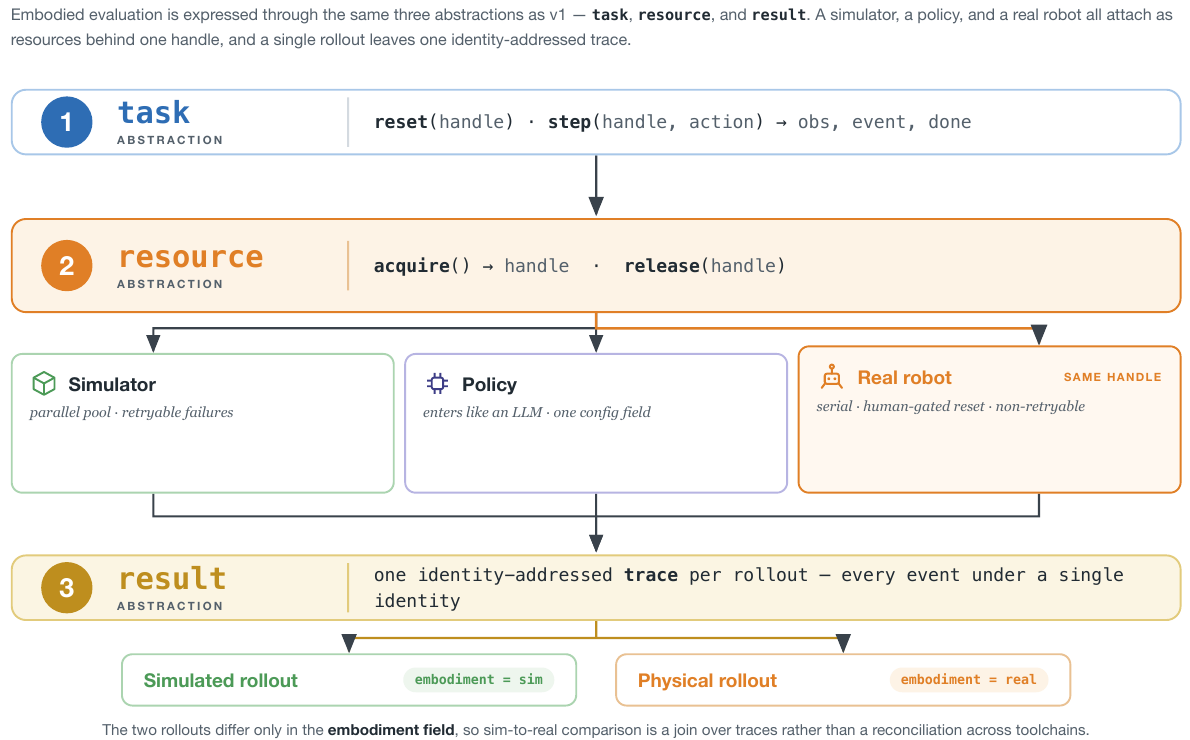}};
    \coordinate (chipTop) at ($(overview.north west)!0.896!(overview.south west)$);
    \coordinate (chipBot) at ($(overview.north west)!0.941!(overview.south west)$);
    \coordinate (chipMid) at ($(overview.north west)!0.918!(overview.south west)$);
    \coordinate (noteTop) at ($(overview.north west)!0.970!(overview.south west)$);
    \coordinate (simL)  at ($(overview.north west)!0.330!(overview.north east)$);
    \coordinate (simR)  at ($(overview.north west)!0.472!(overview.north east)$);
    \coordinate (simC)  at ($(overview.north west)!0.401!(overview.north east)$);
    \coordinate (realL) at ($(overview.north west)!0.740!(overview.north east)$);
    \coordinate (realR) at ($(overview.north west)!0.885!(overview.north east)$);
    \coordinate (realC) at ($(overview.north west)!0.813!(overview.north east)$);
    \fill[white] (noteTop -| overview.south west) rectangle (overview.south east);
    \fill[white] (simL |- chipTop) rectangle (simR |- chipBot);
    \fill[white] (realL |- chipTop) rectangle (realR |- chipBot);
    \node[inner xsep=2pt, inner ysep=0.5pt,
          font=\tiny\ttfamily, text=black!65]
      at (simC |- chipMid)
      {\shortstack{\texttt{execution\_domain =}\\[-1pt]
                   \texttt{simulation}}};
    \node[inner xsep=2pt, inner ysep=0.5pt,
          font=\tiny\ttfamily, text=black!65]
      at (realC |- chipMid)
      {\shortstack{\texttt{execution\_domain =}\\[-1pt]
                   \texttt{real\_robot}}};
  \end{tikzpicture}
  \caption{The embodied evaluation substrate. Task contracts define the
  episode and benchmark semantics; simulators, policies under test, and the
  real robot attach through resource bindings; and every rollout writes into
  an identity-addressed result trace. Matched simulated and physical episodes
  share a parent trace identity while retaining execution-domain-specific
  records.}
  \label{fig:embodied-substrate}
\end{figure}

Bringing embodied evaluation onto the \di substrate---up to and including a
physical robot---is a change of backends, not of architecture. The three v1
invariants enter unchanged: every episode, whatever its shape, is driven
through one \texttt{reset}/\texttt{step} interface with all transient state on
a per-episode handle; every expensive backend is reached only through
\texttt{acquire}/\texttt{release}, its deployment, scaling, and failure
handling staying inside a control plane behind the handle; and every event any
subsystem produces is written into one append-only trace, addressed by a
hierarchical identity that carries causal lineage~\citep{deepinsight}. Nothing
in these invariants says what sits behind a handle. Three new backend classes
exploit that property; \figref{fig:embodied-substrate} sketches how each
attaches.

\paragraph{Simulators enter as sandboxed backends.}
The simulators behind the embodied benchmarks---Habitat, MuJoCo/robosuite,
SAPIEN/ManiSkill---are third-party stacks with mutually conflicting
dependencies. Each ships as a container image behind the sandbox handle,
adopting the containerized view of \citet{vlaevalharness}, so benchmarks stay
dependency-isolated while the plane's lease lifecycle, parallelism, and
accounting apply to them unchanged.

\paragraph{The policy under test enters as an inference backend.}
A model server exposes an observation-in, action-out interface and registers
into the same service catalog as a language model, so substituting the policy
under test is a configuration change, not an integration. Each benchmark
binding declares the observation and action interface it expects from the
policy server; the declaration is validated when benchmark and policy are
bound, so an interface mismatch fails at configuration time rather than
mid-rollout.

\paragraph{The real robot enters as one more backend---asymmetric, but a
backend nonetheless.}
The invariants extend unchanged: the robot is leased per episode, the lease is
bound to the episode identity, and every event it produces writes into the
shared trace. The physical backend's asymmetries---serial capacity where a
simulator pool is parallel, resets that may be human-gated, and interventions
and emergency stops that are physical outcomes rather than retryable
infrastructure errors---are absorbed as backend machinery behind the handle.
Concretely: a safety interlock sits in front of the command stream;
human-gated resets are logged as trace events; interventions and emergency
stops are recorded as first-class outcomes rather than out-of-band incidents;
and binding each episode to a specific robot unit keeps hardware variance
decomposable. Above the handle, a real-robot episode is acquired, driven, and
released like any other.

\paragraph{One trace, two execution domains.}
On the trace, a matched pair of rollouts is organized under a single
\emph{parent trace identity}: the simulated and physical episodes hang as
sibling branches beneath it, each retaining its own execution-domain-specific
record and discriminated by an \texttt{execution\_domain} field
(\texttt{simulation} or \texttt{real\_robot}). What makes the pair
comparable is recorded rather than assumed: the pairing key refers to a
measured initial-state calibration---the robot's starting configuration is
measured, the simulator is initialized from those measurements, and any
residual misalignment is retained as a covariate of the pair. The matched
rollouts share the same task and metric contract while retaining
execution-domain-specific episode records, so cross-domain reduction is a join
on the parent identity rather than a reconciliation across result formats.

The result is that ``the same evaluation, in simulation and on hardware'' is a
well-defined phrase on this substrate: one driver, one evaluation contract,
and one hierarchical trace.

%% file: main/experiments.tex

\section{System 1: Benchmark Fidelity and Common-Protocol Evaluation}
\label{sec:system1}

This section evaluates released \sone checkpoints through the shared \di
runtime. Benchmark-fidelity experiments retain the source task split, reset
distribution, success predicate, observation/action interface, and aggregation
rule. Where a controlled cross-model comparison requires an aligned setting,
the shared configuration and its differences from the source protocols are
stated explicitly. A published value is used as a reference (Ref.) only when
it can be tied to an identified checkpoint and protocol revision; DI denotes
our execution. When a source publishes only aggregate values, the comparison
is aggregate rather than episode-paired. Models without a verifiable reference
are reported as DI-only.

\tabref{tab:s1-coverage} maps the coverage: two navigation benchmarks
(\secref{sec:system1-nav}) and four manipulation benchmarks
(\secref{sec:system1-man}), each entering through the shared driver and
resource bindings of \secref{sec:method}, with released checkpoints as the
systems under test. The remainder of the section reports the resulting
reference comparisons.

\begin{table}[t]
\centering
\scriptsize
\setlength{\tabcolsep}{4pt}
\renewcommand{\arraystretch}{1.2}
\caption{\sone coverage: six benchmarks, two domains, one substrate. Models
are released checkpoints; reported metrics combine benchmark scores and
trace-derived execution diagnostics and are
reported in \tabref{tab:nav-vln}--\tabref{tab:man-robocasa365}.}
\label{tab:s1-coverage}
\begin{tabularx}{\linewidth}{llp{0.16\linewidth}Xl}
\toprule
Domain & Benchmark & Simulator & Models (released checkpoints) & Reported metrics \\
\midrule
\multirow{2}{*}{Navigation}
 & R2R-CE & Habitat-Sim & StreamVLN, RynnBrain-Nav, JanusVLN & SR, SPL, OSR, NE \\
 & RxR-CE & Habitat-Sim & StreamVLN, RynnBrain-Nav, JanusVLN & NE, SR, SPL, nDTW \\
\midrule
\multirow{4}{*}{Manipulation}
 & LIBERO & MuJoCo/robosuite & $\pi_0$, $\pi_{0.5}$, $\pi_0$-FAST, OpenVLA, OpenVLA-OFT & per-suite SR \\
 & LIBERO-Plus & MuJoCo/robosuite & same checkpoints as LIBERO & per-perturbation SR \\
 & SimplerEnv (WidowX/Bridge) & SAPIEN/ManiSkill & StarVLA (3 heads) & SR \\
 & RoboCasa365 & MuJoCo/robosuite & Diffusion Policy, $\pi_0$, $\pi_{0.5}$, GR00T N1.5 & split-level SR \\
\bottomrule
\end{tabularx}
\end{table}

\subsection{Navigation}
\label{sec:system1-nav}

\begin{figure}[ht]
\centering
\includegraphics[width=\linewidth]{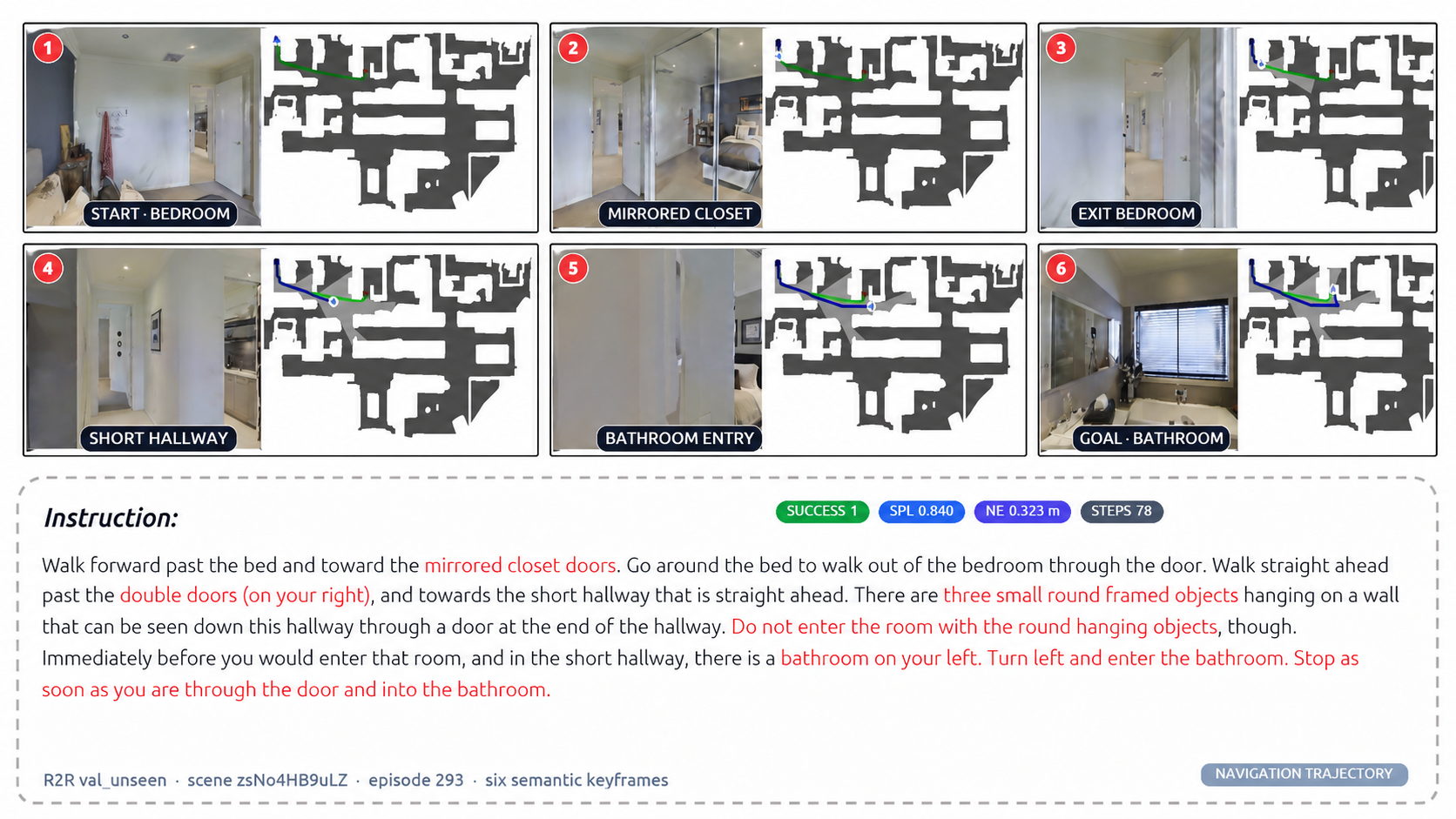}
\caption{Example R2R-CE episode from the unified navigation evaluation. Six
semantic keyframes show the egocentric monocular observations used for policy
inference alongside the trace-derived trajectory visualization, instruction,
and terminal metrics. The top-down map is used only for visualization and is
not exposed to the policy.}
\label{fig:nav-trajectory}
\end{figure}

\paragraph{Unified evaluation configuration.}
We evaluate R2R-CE and the English RxR-CE val-unseen split in Habitat-Sim using
1,839 and 3,669 episodes, respectively. Every DI run uses the same episode list,
an egocentric monocular camera as the policy input, and a maximum horizon of
500 navigation steps. Privileged simulator state is used only by the
environment and metric implementation and is never exposed to the policy. The
models share the same \di episode driver, resource binding, and trace schema;
only checkpoint-specific preprocessing and action decoding remain inside the
model adapter. R2R-CE emphasizes English instruction following and stopping in
continuous scenes, whereas RxR-CE contains longer routes and denser
instruction--trajectory grounding~\citep{krantz2020vlnce,ku2020rxr,
hong2022discretecontinuous}. Figure~\ref{fig:nav-trajectory} illustrates one
R2R-CE episode under this configuration.

The evaluated cohort consists of the released StreamVLN, RynnBrain-Nav, and
JanusVLN checkpoints~\citep{wei2026streamvln,dang2026rynnbrain,
zeng2026janusvln}. The Ref.\ columns provide the values reported by the source
publications, while all DI columns use the same aligned 500-step configuration
and form the controlled cross-model comparison.

\paragraph{Metrics.}
Success rate (SR) measures correct task completion, while success weighted by
path length (SPL) discounts successful but inefficient trajectories
~\citep{anderson2018evaluation}. Navigation error (NE) is the final geodesic
distance to the goal. R2R-CE additionally reports oracle success rate (OSR),
which credits an episode if any visited state enters the success region.
RxR-CE instead reports normalized Dynamic Time Warping (nDTW), which measures
alignment between the executed and reference trajectories
~\citep{ilharco2019dtw}. Bounded metrics are reported on a $0$--$100$ scale;
NE is in meters.

\begin{table}[t]
\centering
\scriptsize
\setlength{\tabcolsep}{5pt}
\renewcommand{\arraystretch}{1.15}
\caption{Navigation results on val-unseen splits. All runs use egocentric
monocular input, and all DI values use a maximum of 500 steps. For each
metric, Ref.\ (gray) is the value reported by the corresponding source
publication and DI is our unified execution. Bounded metrics are percentages;
NE is in meters.}
\label{tab:nav-vln}

\textbf{(a) R2R-CE, 1,839 episodes.}\par\smallskip
\begin{tabular}{l cc cc cc cc}
\toprule
\multirow{2}{*}[-2pt]{Model}
& \multicolumn{2}{c}{SR$\uparrow$}
& \multicolumn{2}{c}{SPL$\uparrow$}
& \multicolumn{2}{c}{OSR$\uparrow$}
& \multicolumn{2}{c}{NE$\downarrow$} \\
\cmidrule(lr){2-3}\cmidrule(lr){4-5}\cmidrule(lr){6-7}\cmidrule(lr){8-9}
& Ref. & DI & Ref. & DI & Ref. & DI & Ref. & DI \\
\midrule
StreamVLN~\citep{wei2026streamvln}
 & \refv{56.40} & 55.85 & \refv{50.20} & 49.97
 & \refv{63.60} & 63.13 & \refv{4.90} & 5.08 \\
RynnBrain-Nav~\citep{dang2026rynnbrain}
 & \refv{58.60} & 57.31 & \refv{49.60} & 48.01
 & \refv{71.60} & 69.55 & \refv{4.92} & 5.02 \\
JanusVLN~\citep{zeng2026janusvln}
 & \refv{60.50} & 57.26 & \refv{56.80} & 52.99
 & \refv{65.20} & 62.91 & \refv{4.78} & 4.92 \\
\bottomrule
\end{tabular}

\vspace{7pt}
\textbf{(b) RxR-CE English, 3,669 episodes.}\par\smallskip
\begin{tabular}{l cc cc cc cc}
\toprule
\multirow{2}{*}[-2pt]{Model}
& \multicolumn{2}{c}{SR$\uparrow$}
& \multicolumn{2}{c}{SPL$\uparrow$}
& \multicolumn{2}{c}{nDTW$\uparrow$}
& \multicolumn{2}{c}{NE$\downarrow$} \\
\cmidrule(lr){2-3}\cmidrule(lr){4-5}\cmidrule(lr){6-7}\cmidrule(lr){8-9}
& Ref. & DI & Ref. & DI & Ref. & DI & Ref. & DI \\
\midrule
StreamVLN~\citep{wei2026streamvln}
 & \refv{54.40} & 54.40 & \refv{45.40} & 45.81
 & \refv{63.70} & 64.76 & \refv{5.65} & 5.60 \\
RynnBrain-Nav~\citep{dang2026rynnbrain}
 & \refv{56.10} & 56.85 & \refv{42.70} & 43.97
 & \refv{59.60} & 59.20 & \refv{6.20} & 5.57 \\
JanusVLN~\citep{zeng2026janusvln}
 & \refv{56.20} & 55.36 & \refv{47.50} & 46.56
 & \refv{62.10} & 61.62 & \refv{6.06} & 5.69 \\
\bottomrule
\end{tabular}
\end{table}

\paragraph{Reproduction fidelity and residual differences.}
Overall, the DI results remain close to the published references and preserve
their metric scale and qualitative behavior. On R2R-CE, StreamVLN differs by
at most 0.55 percentage points across the bounded metrics, RynnBrain-Nav by
2.05 points, and JanusVLN by 3.81 points; their NE differences are 0.18,
0.10, and 0.14 meters, respectively. The RxR-CE reproduction is similarly
close: StreamVLN matches the reported SR exactly, while the bounded-metric
differences for all three checkpoints are at most 1.27 points and the NE
differences are at most 0.63 meters. These results indicate that the shared
runtime largely preserves the readings of the original evaluation stacks.

The remaining gaps are consistent with several protocol- and
implementation-level effects. First, the source publications provide aggregate
results rather than episode-level outputs with matched random seeds, so
Ref.--DI comparisons cannot control simulator stochasticity and episode
ordering exactly. Second, residual differences in software revisions and at
the adapter boundary---including image preprocessing, action decoding,
collision handling, and stopping implementation---can change an early motion
and then compound through closed-loop execution. This accumulation is
especially visible on the longer RxR-CE trajectories, where modest per-step
deviations can alter both endpoint metrics and nDTW without indicating a
systematic failure of the reproduced policy. All DI results nevertheless use
the same episode lists, monocular observations, and 500-step limit, so these
residual gaps are not caused by unequal evaluation budgets.
\FloatBarrier

\subsection{Manipulation}
\label{sec:system1-man}

\paragraph{Benchmark coverage.}
We ask whether one evaluation substrate can retain native benchmark behavior
across changes in simulator, embodiment, and task structure. The selected
benchmarks provide complementary tests: LIBERO supplies a clean-suite baseline
for single-arm manipulation; LIBERO-Plus adds seven controlled perturbation
families; SimplerEnv moves the evaluation to SAPIEN/ManiSkill and the
WidowX/Bridge embodiment; and RoboCasa365 introduces mobile manipulation and
compositional household tasks
\citep{liu2023libero,fei2026liberoplus,li2024simplerenv,robocasa365}.
Together, they separate clean-suite execution, perturbation sensitivity,
cross-backend and cross-embodiment portability, and compositional task
evaluation.

\paragraph{Evaluation setup.}
We evaluate identifiable, author-released or benchmark-distributed checkpoints
from the $\pi$, OpenVLA, StarVLA, Diffusion Policy, and GR00T families
\citep{black2024pi0,physicalintelligence2025pi05,pertsch2025fast,
kim2024openvla,kim2025openvlaoft,starvla2026,chi2023diffusionpolicy,
nvidia2025grootn15}. Each DI run retains the benchmark's task split, reset
distribution, observation and action interfaces, success criterion, and
aggregation rule. LIBERO references come from the corresponding model
releases, with OpenVLA-OFT using the official 4-in-1 checkpoint
~\citep{kim2025openvlaoftcombined}. LIBERO-Plus reuses these checkpoints
without post-training; because it provides no $\pi_{0.5}$ reference, we use an
independent study of the official checkpoint~\citep{zhang2026wamrobustness}.
SimplerEnv references are the StarVLA model-zoo aggregates. StarVLA-OFT is
excluded because the released Bridge checkpoint was confirmed to be incorrect
and no corrected artifact was available~\citep{starvlaoftcheckpoint2026}.
RoboCasa365 uses official leaderboard references; all DI runs use v1.0.1, and
an effective-step audit verifies that the official $1.5\times$ horizon is
applied once and aligned with the reference evaluations
~\citep{robocasa365code,robocasa365leaderboard}.

\paragraph{Evaluation scale.}
LIBERO runs $4$ suites $\times$ $10$ tasks $\times$ $50$ episodes, totaling
2,000 episodes per model; LIBERO-Plus evaluates 10,030 test instances once
each. SimplerEnv runs four sweeps over the 24 official object-pose
configurations for each of four WidowX tasks, totaling 384 episodes per model
~\citep{li2024simplerenv,starvla2026}; its reported average is the unweighted
mean of the four task rates. RoboCasa365 runs 50 episodes on each of 50 tasks,
totaling 2,500 episodes per model, split into 900 atomic-seen, 800
composite-seen, and 800 composite-unseen episodes.

\FloatBarrier

\begin{table}[!htbp]
\centering
\scriptsize
\setlength{\tabcolsep}{4pt}
\renewcommand{\arraystretch}{1.15}
\caption{LIBERO success rates across the four standard suites. For each suite,
Ref.\ (gray) is the published value and DI is our execution.}
\label{tab:man-libero}
\begin{tabular}{l cc cc cc cc cc}
\toprule
\multirow{2}{*}[-2pt]{Model}
& \multicolumn{2}{c}{Spatial$\uparrow$}
& \multicolumn{2}{c}{Object$\uparrow$}
& \multicolumn{2}{c}{Goal$\uparrow$}
& \multicolumn{2}{c}{LIBERO-10$\uparrow$}
& \multicolumn{2}{c}{Avg.$\uparrow$} \\
\cmidrule(lr){2-3}\cmidrule(lr){4-5}\cmidrule(lr){6-7}
\cmidrule(lr){8-9}\cmidrule(lr){10-11}
& Ref. & DI & Ref. & DI & Ref. & DI & Ref. & DI & Ref. & DI \\
\midrule
$\pi_0$
 & \refv{96.80} & 97.80 & \refv{98.80} & 97.40 & \refv{95.80} & 92.80
 & \refv{85.20} & 81.40 & \refv{94.15} & 92.35 \\
$\pi_{0.5}$
 & \refv{98.80} & 98.80 & \refv{98.20} & 98.40 & \refv{98.00} & 97.20
 & \refv{92.40} & 91.80 & \refv{96.85} & 96.55 \\
$\pi_0$-FAST
 & \refv{96.40} & 97.00 & \refv{96.80} & 97.20 & \refv{88.60} & 87.20
 & \refv{60.20} & 62.20 & \refv{85.50} & 85.90 \\
OpenVLA
 & \refv{84.70} & 86.20 & \refv{88.40} & 86.80 & \refv{79.20} & 77.80
 & \refv{53.70} & 54.00 & \refv{76.50} & 76.20 \\
OpenVLA-OFT (comb.)
 & \refv{96.80} & 98.00 & \refv{97.00} & 97.80 & \refv{95.40} & 96.40
 & \refv{98.00} & 95.40 & \refv{96.80} & 96.90 \\
\bottomrule
\end{tabular}
\end{table}

\begin{table}[!htbp]
\centering
\scriptsize
\setlength{\tabcolsep}{2.5pt}
\renewcommand{\arraystretch}{1.15}
\caption{LIBERO-Plus success rates by perturbation family. Total is
episode-weighted; Ref.\ (gray) is the published value and DI is our
execution. $\dagger$: the published $\pi_0$ Sensor-Noise reference is an
unresolved benchmark-reference issue (see text).}
\label{tab:man-liberoplus}
\begin{tabular}{l cc cc cc cc cc cc cc cc}
\toprule
\multirow{2}{*}[-2pt]{Model}
& \multicolumn{2}{c}{Camera}
& \multicolumn{2}{c}{Robot init.}
& \multicolumn{2}{c}{Lang.}
& \multicolumn{2}{c}{Light}
& \multicolumn{2}{c}{Backgr.}
& \multicolumn{2}{c}{Noise}
& \multicolumn{2}{c}{Layout}
& \multicolumn{2}{c}{Total$\uparrow$} \\
\cmidrule(lr){2-3}\cmidrule(lr){4-5}\cmidrule(lr){6-7}\cmidrule(lr){8-9}
\cmidrule(lr){10-11}\cmidrule(lr){12-13}\cmidrule(lr){14-15}
\cmidrule(lr){16-17}
& Ref. & DI & Ref. & DI & Ref. & DI & Ref. & DI
& Ref. & DI & Ref. & DI & Ref. & DI & Ref. & DI \\
\midrule
$\pi_0$
 & \refv{13.8} & 17.0 & \refv{6.0} & 7.5 & \refv{58.8} & 64.9
 & \refv{85.0} & 86.9 & \refv{81.4} & 83.3
 & \refv{79.0}\rlap{$^\dagger$} & 18.7 & \refv{68.9} & 73.3
 & \refv{53.6} & 46.8 \\
$\pi_{0.5}$
 & \refv{75.4} & 71.9 & \refv{77.5} & 74.0 & \refv{85.6} & 85.6
 & \refv{96.9} & 96.7 & \refv{94.6} & 95.0 & \refv{89.7} & 86.8
 & \refv{85.7} & 86.7 & \refv{85.7} & 84.3 \\
$\pi_0$-FAST
 & \refv{65.1} & 63.1 & \refv{21.6} & 22.8 & \refv{61.0} & 69.5
 & \refv{73.2} & 75.6 & \refv{73.2} & 74.5 & \refv{74.4} & 72.0
 & \refv{68.8} & 73.8 & \refv{61.6} & 63.6 \\
OpenVLA
 & \refv{0.8} & 0.5 & \refv{3.5} & 3.7 & \refv{23.0} & 37.5
 & \refv{8.1} & 13.7 & \refv{34.8} & 36.3 & \refv{15.2} & 5.9
 & \refv{28.5} & 42.2 & \refv{15.6} & 19.2 \\
OpenVLA-OFT (comb.)
 & \refv{55.6} & 55.9 & \refv{21.7} & 24.1 & \refv{81.0} & 85.2
 & \refv{92.7} & 93.3 & \refv{91.0} & 87.4 & \refv{78.6} & 73.8
 & \refv{68.7} & 74.0 & \refv{67.9} & 68.7 \\
\bottomrule
\end{tabular}
\end{table}

\begin{table}[!htbp]
\centering
\scriptsize
\setlength{\tabcolsep}{5pt}
\renewcommand{\arraystretch}{1.15}
\caption{SimplerEnv WidowX/Bridge Visual-Matching success rates. The model
zoo publishes only the four-task average, so per-task values are DI-only;
for the average, Ref.\ (gray) is the published value and DI is our
execution.}
\label{tab:man-simpler}
\begin{tabular}{l cccc cc}
\toprule
\multirow{2}{*}[-2pt]{Model}
& \multicolumn{4}{c}{Per-task success (DI)}
& \multicolumn{2}{c}{Avg.$\uparrow$} \\
\cmidrule(lr){2-5}\cmidrule(lr){6-7}
& Stack cube & Carrot/Plate & Spoon/Towel & Eggplant/Basket & Ref. & DI \\
\midrule
StarVLA-FAST  & 41.7 & 45.8 & 75.0 & 87.5 & \refv{58.6} & 62.5 \\
StarVLA-$\pi$ & 34.4 & 69.8 & 82.3 & 76.0 & \refv{62.5} & 65.6 \\
StarVLA-GR00T & 40.6 & 55.2 & 81.3 & 77.1 & \refv{63.6} & 63.5 \\
\bottomrule
\end{tabular}
\end{table}

\begin{table}[!htbp]
\centering
\scriptsize
\setlength{\tabcolsep}{5pt}
\renewcommand{\arraystretch}{1.15}
\caption{RoboCasa365 multi-task-learning success rates by task split. For
each split, Ref.\ (gray) is the official leaderboard value and DI is our
execution.}
\label{tab:man-robocasa365}
\begin{tabular}{l cc cc cc cc}
\toprule
\multirow{2}{*}[-2pt]{Model}
& \multicolumn{2}{c}{Atomic seen$\uparrow$}
& \multicolumn{2}{c}{Comp. seen$\uparrow$}
& \multicolumn{2}{c}{Comp. unseen$\uparrow$}
& \multicolumn{2}{c}{Overall$\uparrow$} \\
\cmidrule(lr){2-3}\cmidrule(lr){4-5}\cmidrule(lr){6-7}\cmidrule(lr){8-9}
& Ref. & DI & Ref. & DI & Ref. & DI & Ref. & DI \\
\midrule
Diffusion Policy
 & \refv{15.70} & 16.56 & \refv{0.20} & 0.13
 & \refv{1.25} & 0.13 & \refv{6.10} & 6.04 \\
$\pi_0$
 & \refv{34.60} & 34.67 & \refv{6.10} & 5.88
 & \refv{1.10} & 1.63 & \refv{14.80} & 14.88 \\
$\pi_{0.5}$
 & \refv{39.60} & 41.67 & \refv{7.10} & 6.13
 & \refv{1.20} & 2.13 & \refv{16.90} & 17.64 \\
GR00T N1.5
 & \refv{50.70} & 49.00 & \refv{14.80} & 16.25
 & \refv{2.70} & 4.13 & \refv{23.90} & 24.16 \\
\bottomrule
\end{tabular}
\end{table}

\begin{figure}[!htbp]
\centering
\includegraphics[width=\linewidth]{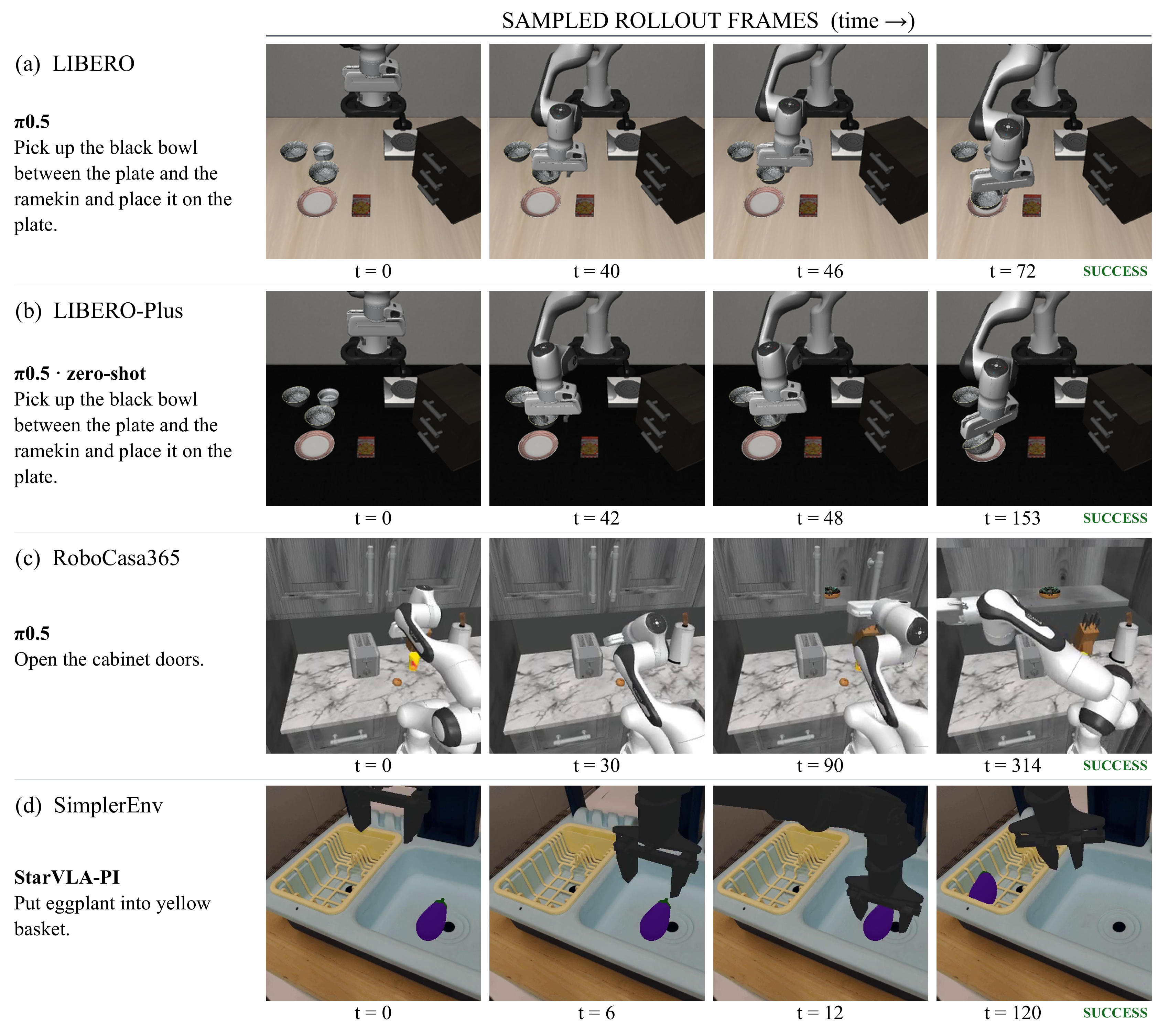}
\caption{Representative successful closed-loop rollouts from LIBERO,
LIBERO-Plus, RoboCasa365, and SimplerEnv (top to bottom). Each row identifies
the evaluated checkpoint and task and shows the reset observation ($t=0$), two
intermediate observations, and the final sampled observation at the indicated
saved-rollout frame indices. Intervening observation--action steps are omitted;
all four episodes satisfy their benchmark's native success predicate.}
\label{fig:manipulation-rollouts}
\end{figure}

\paragraph{Qualitative rollout evidence.}
The aggregate scores above reduce each episode to a terminal success value,
whereas \figref{fig:manipulation-rollouts} exposes the trajectory-level
interaction that produces that outcome. Across four representative successful
cases, the observations evolve as policy actions modify the scene over
benchmark-specific horizons. This dependence distinguishes VLA evaluation from
single-turn VLM scoring: later policy inputs are consequences of earlier
actions within the same environment-coupled rollout. The displayed frames are
sparse samples and do not imply a shared phase structure across tasks.

\FloatBarrier

\paragraph{Overall results.}
Across standard LIBERO, SimplerEnv, and RoboCasa365, the DI aggregates track
their published references without a systematic upward or downward shift
across model families. Most LIBERO-Plus model--perturbation pairs follow the
same pattern. The material exception is $\pi_0$ under Sensor Noise: DI obtains
18.7, compared with the published reference of 79.0. This category
accounts for most of the $\pi_0$ Total difference. Independent evaluations
using the public OpenPI checkpoint report the same 16--20\% range, while the
maintainers are auditing the provenance of the original reference result
~\citep{liberoplusnoiseissue2026}. We therefore treat this entry (marked
$\dagger$ in \tabref{tab:man-liberoplus}) as an unresolved
benchmark-reference issue and exclude it from behavioral and
model-ranking claims. Taken together, the quantitative comparisons and
trajectory examples support the intended benchmark-level fidelity claim: the
shared runtime does not introduce a systematic shift across the evaluated
simulators, embodiments, and checkpoints, while localized discrepancies remain
attributable to specific benchmark--reference pairs.

\section{System 0: Scalable WBC Evaluation from Simulation to Real-Robot Evidence}
\label{sec:system0}

V1 used trace-backed simulation to screen and diagnose internal \szero
policies~\citep[\S5.2]{deepinsight}. This section turns that case study into a
standardized evaluation and extends it in two directions: MotionBench freezes
a common workload and metric contract across released WBC models, and matched
real-robot execution turns simulation-only evidence into measured
simulation--real correspondence. The frozen contract matters because WBC
evaluation varies along three coupled axes---the motion workload, the WBC
interface, and the execution domain---so evaluating each model through its own
harness confounds model behavior with task semantics, interface conversion,
and reduction rules. Here, each released model instead attaches through a
versioned binding, scaling the same evaluation across architectures,
checkpoints, and runtime interfaces while the workload, trace schema, and
metric reducers stay fixed. \secref{sec:system0-bench} fixes the workload
and metrics; \secref{sec:system0-sim} compares four released WBC models and
qualifies hardware cases; and \secref{sec:system0-real} evaluates the two
SONIC variants in matched real-robot trials.

\begin{figure}[H]
\centering
\includegraphics[width=\linewidth]{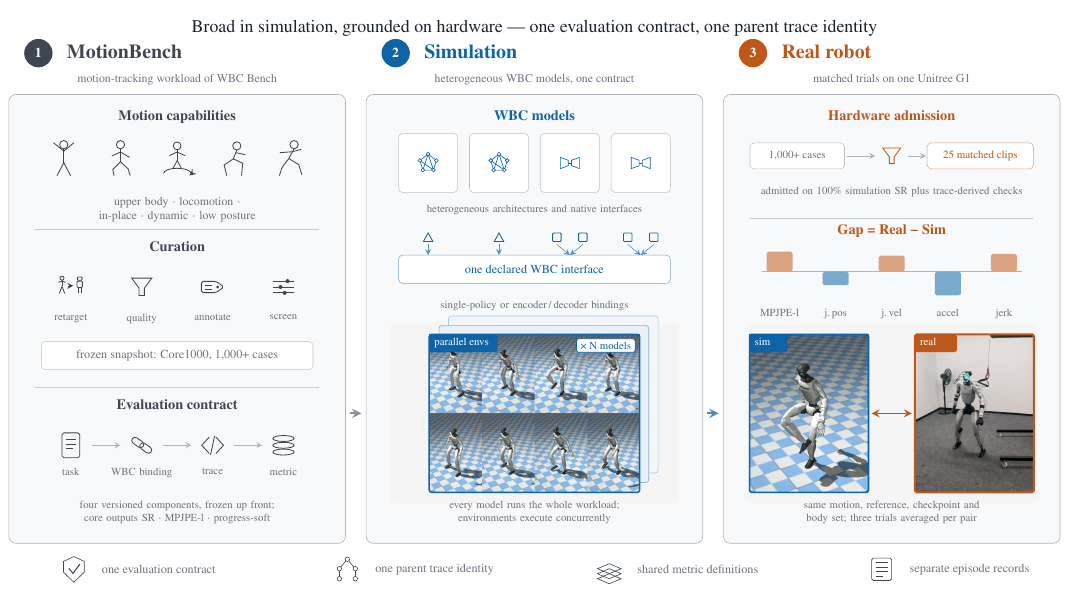}
\caption{System~0 evaluation overview. MotionBench fixes the versioned
workload and measurement semantics; heterogeneous WBC models are compared in
a common simulation space; and trace-qualified WBC--motion pairs enter
matched real-robot trials for simulation--reality analysis. SR is used only
in simulation; hardware correspondence uses matched root-relative and
joint-space measurements.}
\label{fig:system0-overview}
\end{figure}

\subsection{Standardized WBC Evaluation}
\label{sec:system0-bench}

\emph{MotionBench} is the motion-tracking workload of our internally developed
WBC Bench. Here it provides a curated, versioned \szero workload for evaluating
released WBC models. Candidate references are retargeted to the evaluation
robot profile, quality-assessed, manually annotated, and screened by difficulty
and motion type. Each frozen snapshot assigns a stable identity and metadata
to every case. For report-level aggregation, cases are grouped into five
capabilities---upper body, locomotion, in-place, dynamic, and low posture---and
can also be decomposed by duration and motion.

The common evaluation contract is instantiated through four versioned
components: the task component fixes the case and reference; the WBC binding
component declares the model interface; the trace component specifies retained
evidence; and the metric component fixes its interpretation. Replacing a WBC
model changes only its registry record and binding; the workload, failure
semantics, trace schema, and reduction remain fixed.

The headline simulation metrics separate executability from fidelity.
Throughout this section, SR is the fraction of motions that avoid a hard
failure: root-height or end-effector-height error above 0.25~m, or
root-orientation error above 1.0~rad~\citep{luo2026sonic}. MPJPE-l is
root-translation-subtracted mean per-body position error without rotational
alignment; it is stored in millimeters and reported in centimeters.
Progress-soft is the macro-average of each motion's completion ratio before a
hard failure or global position/heading drift.
Together with the workload and versioned contracts above, these metrics define
the System~0 evaluation specification summarized in
\tabref{tab:wbc-bench-spec}.

\begin{table}[!htbp]
\centering
\scriptsize
\setlength{\tabcolsep}{4pt}
\renewcommand{\arraystretch}{1.2}
\caption{MotionBench specification. The benchmark fixes the workload and
evaluation semantics before model-specific execution.}
\label{tab:wbc-bench-spec}
\begin{tabularx}{\linewidth}{p{0.19\linewidth}X p{0.30\linewidth}}
\toprule
Component & Specification & Evaluation role \\
\midrule
Snapshot & Frozen MotionBench snapshot (Core1000; 1,000+ cases) & Stable case-level comparison \\
Curation & Robot-profile retargeting, quality assessment, manual annotation, and difficulty/type screening & Controlled motion workload \\
Coverage & Upper-body, locomotion, in-place, dynamic, and low-posture motion across multiple durations & Capability and duration decomposition \\
WBC binding & Versioned checkpoint and declared WBC interface under the common evaluation contract & Comparable execution across WBC models \\
Core outputs & SR, MPJPE-l, and progress-soft & Executability, local fidelity, and usable completion \\
\bottomrule
\end{tabularx}
\end{table}

\subsection{Unified Simulation Evaluation of Heterogeneous WBC Models}
\label{sec:system0-sim}

We evaluate four released WBC models with heterogeneous architectures
and runtime interfaces on the same MotionBench snapshot. HoloMotion uses a
causal decoder-only Transformer with sparsely activated MoE layers, whereas
SONIC uses an MLP encoder--decoder architecture with an FSQ token
representation~\citep{chen2026holomotion,luo2026sonic}. In our evaluation,
HoloMotion enters through a single-policy binding, whereas SONIC uses separate
encoder and decoder bindings. All models share the workload, failure
semantics, trace schema, and reducers.

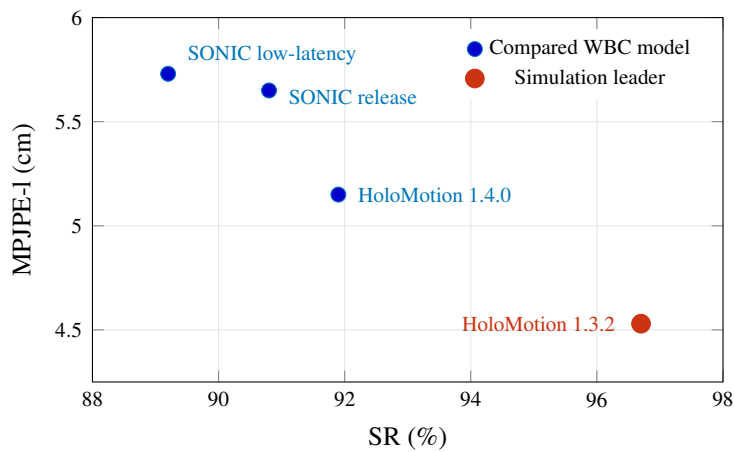
\begin{figure}[!htbp]
\centering
\begin{tikzpicture}
\begin{axis}[
  width=0.62\linewidth,
  height=0.40\linewidth,
  xmin=88, xmax=98,
  ymin=4.25, ymax=6.0,
  xlabel={SR (\%)},
  ylabel={MPJPE-l (cm)},
  grid=major,
  major grid style={black!10},
  tick label style={font=\scriptsize},
  label style={font=\small},
  legend style={font=\scriptsize, draw=none, at={(0.97,0.97)}, anchor=north east},
]
\addplot+[only marks, mark=*, color=PolicyPoint, mark size=2.8pt]
  coordinates {(91.9,5.15) (90.8,5.65) (89.2,5.73)};
\addlegendentry{Compared WBC model}
\addplot+[only marks, mark=*, color=SelectedPolicy, mark size=3.5pt,
  mark options={fill=SelectedPolicy}]
  coordinates {(96.7,4.53)};
\addlegendentry{Simulation leader}
\node[font=\scriptsize, text=SelectedPolicy, anchor=east]
  at (axis cs:96.45,4.53) {HoloMotion 1.3.2};
\node[font=\scriptsize, text=PolicyPoint, anchor=west]
  at (axis cs:92.05,5.15) {HoloMotion 1.4.0};
\node[font=\scriptsize, text=PolicyPoint, anchor=west]
  at (axis cs:90.95,5.62) {SONIC release};
\node[font=\scriptsize, text=PolicyPoint, anchor=west]
  at (axis cs:89.35,5.82) {SONIC low-latency};
\end{axis}
\end{tikzpicture}
\caption{Simulation SR--MPJPE-l plane on MotionBench. Each point is one
released WBC model evaluated under the same task and metric contract.}
\label{fig:wbc-sim-plane}
\end{figure}

\figref{fig:wbc-sim-plane} presents SR and MPJPE-l as complementary measures
of executability and local fidelity, with exact aggregates reported in
\tabref{tab:wbc-sim-metrics}. HoloMotion 1.3.2 leads the full benchmark on both
measures, with 96.7\% SR and 4.53~cm MPJPE-l; it also achieves the highest
progress-soft value (93.1\%).

\begin{table}[!htbp]
\centering
\scriptsize
\setlength{\tabcolsep}{5pt}
\renewcommand{\arraystretch}{1.15}
\caption{MotionBench simulation results. SR and progress-soft are percentages;
MPJPE-l is the macro average over motions.}
\label{tab:wbc-sim-metrics}
\begin{tabular}{l l ccc}
\toprule
WBC & Architecture & SR$\uparrow$ & Progress-soft$\uparrow$ &
MPJPE-l (cm)$\downarrow$ \\
\midrule
HoloMotion 1.3.2 & Sparse-MoE causal Transformer & 96.7 & 93.1 & 4.53 \\
HoloMotion 1.4.0 & Sparse-MoE causal Transformer & 91.9 & 77.9 & 5.15 \\
SONIC release & MLP encoder--decoder + FSQ & 90.8 & 72.7 & 5.65 \\
SONIC low-latency & MLP encoder--decoder + FSQ & 89.2 & 67.4 & 5.73 \\
\bottomrule
\end{tabular}
\end{table}

The capability decomposition in \tabref{tab:wbc-bucket-progress} localizes the
aggregate differences. Upper-body progress-soft is nearly saturated across
models, whereas locomotion, dynamic motion, and low-posture motion produce
larger separation. The SONIC variants achieve hard-failure SRs of 90.8\% and
89.2\%, while their progress-soft values are 72.7\% and 67.4\%, respectively,
revealing a larger separation under a metric that accounts for partial
completion and global drift.
HoloMotion 1.4.0 shows its largest deficits on locomotion and dynamic motion.

\begin{table}[!htbp]
\centering
\scriptsize
\setlength{\tabcolsep}{5pt}
\renewcommand{\arraystretch}{1.15}
\caption{MotionBench progress-soft (\%) by capability. The decomposition
localizes losses in usable tracking behind aggregate SR.}
\label{tab:wbc-bucket-progress}
\begin{tabular}{l ccccc}
\toprule
WBC & Upper body & Locomotion & In-place & Dynamic & Low posture \\
\midrule
HoloMotion 1.3.2 & 97.7 & 95.1 & 91.4 & 90.4 & 82.5 \\
HoloMotion 1.4.0 & 95.2 & 71.0 & 83.1 & 54.3 & 68.0 \\
SONIC release & 93.8 & 63.4 & 74.7 & 55.7 & 55.6 \\
SONIC low-latency & 93.8 & 52.5 & 70.5 & 48.4 & 50.8 \\
\bottomrule
\end{tabular}
\end{table}

For the matched real-robot study, we predeclare SONIC release and SONIC
low-latency as the hardware cohort. They share the same architecture and WBC
interface, enabling a release-level comparison under the same protocol.

\subsection{Grounding Simulation Findings in Real-Robot Evidence}
\label{sec:system0-real}

\paragraph{Hardware qualification.}
Before hardware execution, we freeze a shared subset of 25 motion clips for
the two SONIC variants. A WBC--motion pair is admitted after achieving 100\%
simulation SR and passing trace-derived checks of trajectory completion,
local tracking, root and end-effector posture, and actuator/contact behavior.
Qualified pairs are executed three times on one Unitree G1 from the same
initialization protocol. Simulation SR serves as the qualification metric;
hardware evaluation uses root-relative pose and proprioceptive joint-space
measurements.

\paragraph{Matched protocol.}
Each simulation rollout and its three hardware trials share a parent trace
identity and bind the same motion, reference, checkpoint, adapter version,
body set, and metric definitions, while retaining separate episode records.
Hardware MPJPE-l is computed from synchronized robot-state estimates and
forward kinematics using the same root-translation subtraction as in
simulation. Joint-position and joint-velocity RMSE use the matched reference
trajectory; acceleration and jerk statistics use the executed joint
trajectories. The three hardware trials are averaged for each WBC--motion pair,
followed by an average over matched motions.

\paragraph{Simulation--real correspondence.}
\tabref{tab:wbc-real-correspondence} reports the matched kinematic and dynamic
profile of both SONIC variants. MPJPE-l measures root-relative pose fidelity;
joint-position and joint-velocity RMSE measure reference tracking; and
acceleration and jerk RMS/P95 summarize sustained and upper-tail joint
dynamics. The simulation entries are recomputed on the same 25-clip subset and
therefore differ from the full-MotionBench aggregates in
\tabref{tab:wbc-sim-metrics}.

\begin{table}[!htbp]
\centering
\scriptsize
\setlength{\tabcolsep}{4pt}
\renewcommand{\arraystretch}{1.15}
\caption{Matched simulation--real kinematic and dynamic profile for the
qualified SONIC cohort over 25 matched motion clips. For each variant, Sim.\
(gray) and Real give the paired aggregates and Gap is
$(\mathrm{Real}-\mathrm{Sim})/\mathrm{Sim}\times100\%$. Hardware values
average three trials per WBC--motion pair before aggregation across motions.}
\label{tab:wbc-real-correspondence}
\begin{tabularx}{\linewidth}{l ccc ccc X}
\toprule
\multirow{2}{*}[-2pt]{Metric}
& \multicolumn{3}{c}{SONIC release}
& \multicolumn{3}{c}{SONIC low-latency}
& \multirow{2}{*}[-2pt]{Diagnostic role} \\
\cmidrule(lr){2-4}\cmidrule(lr){5-7}
& Sim. & Real & Gap (\%) & Sim. & Real & Gap (\%) & \\
\midrule
MPJPE-l (cm)
 & \refv{3.08} & 3.41 & +10.83 & \refv{3.28} & 3.89 & +18.65
 & Root-local whole-body tracking \\
Joint-position RMSE (rad)
 & \refv{0.13} & 0.16 & +22.15 & \refv{0.15} & 0.18 & +22.53
 & Joint pose tracking and steady bias \\
Joint-velocity RMSE (rad/s)
 & \refv{0.74} & 0.75 & +1.58 & \refv{0.80} & 0.88 & +10.46
 & Dynamic tracking, latency, and damping \\
Joint-acceleration RMS (rad/s$^2$)
 & \refv{18.40} & 14.16 & $-$23.05 & \refv{22.32} & 20.39 & $-$8.66
 & Sustained high-frequency activity \\
Joint-jerk RMS (rad/s$^3$)
 & \refv{1181.51} & 941.36 & $-$20.33 & \refv{1442.24} & 1375.07 & $-$4.66
 & Sustained command discontinuity \\
Joint-acceleration P95 (rad/s$^2$)
 & \refv{27.01} & 27.69 & +2.52 & \refv{32.95} & 39.07 & +18.58
 & Transient acceleration peaks \\
Joint-jerk P95 (rad/s$^3$)
 & \refv{1518.76} & 1770.80 & +16.60 & \refv{1854.33} & 2517.01 & +35.74
 & Sharp spikes and actuator stress \\
\bottomrule
\end{tabularx}
\end{table}

SONIC release remains lower than SONIC low-latency across all seven readings
in both domains. Their joint-position gaps are similar, whereas low-latency
shows larger MPJPE-l (+18.65\% versus +10.83\%), joint-velocity (+10.46\%
versus +1.58\%), acceleration-P95 (+18.58\% versus +2.52\%), and jerk-P95
(+35.74\% versus +16.60\%) gaps. For both variants, hardware increases
tracking errors and tail peaks while reducing acceleration/jerk RMS.
Acceleration and jerk gaps characterize motion dynamics rather than
independent success criteria.

\FloatBarrier

\section{Full-System: Handoff Attribution and Repairability in Physical
Execution}
\label{sec:full-system}

V1 localized composed \ssystem{2--1--0} failures to the task, subgoal, and
subsystem levels over the shared trace~\citep{deepinsight}. \di II adds a
second automatic stage: when the localized failure falls on a system handoff,
attribution identifies the specific cross-module assumption, interface
convention, or coordination design that broke, and maps it to a repair
action. Every decision must be supported by a concrete trace interval. We
evaluate the mechanism in simulation and on a real robot. Attribution sorts
handoff failures into two causes whose repairs cannot substitute for each
other:

\begin{itemize}\setlength\itemsep{0.15em}
  \item \textbf{Boundary-definition problem.} The upstream termination
  criterion and the downstream precondition are not defined over the same
  state space, the two criteria evaluate different quantities, or a necessary
  condition was never registered in the interface. Adding the missing interface
  convention repairs it.

  \item \textbf{Executability problem.} The boundary is correct, but the
  downstream feasible entry region is comparable in width to the minimum
  executable step, so corrections cannot land inside it reliably---and cannot at
  all once the region is narrower than one step. No interface convention repairs
  this; the action space must change.
\end{itemize}

Separating the two converts \emph{handoff failure} from an undifferentiated
label into an actionable repair direction.

\subsection{Automatic Attribution of Handoff Failures}
\label{sec:full-system-diagnosis}

We reuse the object-introduction task of the first report---find and greet the
user, resolve a referential introduction request, take up a pose from which the
object (here a vehicle) can be presented, present it, and terminate---whose
chain contains six handoffs, H1--H6
(\tabref{tab:fs-chain}). Each skill is
first run from controlled initial states to register its
contract---termination signal, start condition, observable
postcondition---making implicit handoff assumptions testable. Analysis
concentrates on the alignment handoffs H2 (navigation~$\to$~greeting) and H5
(navigation~$\to$~presentation), whose geometric preconditions admit a
quantitative repairability judgment.

Attribution checks each state declared at the localized boundary against an
independently reconstructed one---rules evaluate geometric and controller
predicates, a multimodal judge the semantic and visual ones---and yields five
labels, each mapped to one repair action (\tabref{tab:fs-labels}); when
evidence is insufficient, attribution abstains rather than forcing a label.
The declared--reconstructed split is not redundancy: embodied agents' terminal
self-reports are known to diverge from their verified world
state~\citep{chen2026vigil}, and a termination error is that divergence
surfacing at a handoff boundary.
Simulation additionally records privileged pose, collision state, and
multi-view video around each handoff, used only to supervise and audit the
diagnosis.

\begin{table}[!htbp]
\centering
\scriptsize
\setlength{\tabcolsep}{4pt}
\renewcommand{\arraystretch}{1.2}
\caption{The five attribution labels: each verified evidence pattern maps to
one repair action. The first three apply at the handoff boundary, the last two
inside the downstream skill's execution; abbreviations are used in
\tabref{tab:fs-chain}.}
\label{tab:fs-labels}
\begin{tabularx}{\linewidth}{>{\raggedright\arraybackslash}X
                             >{\raggedright\arraybackslash}p{0.185\linewidth}
                             >{\raggedright\arraybackslash}p{0.24\linewidth}}
\toprule
Evidence pattern & Label & Repair \\
\midrule
Source reports success; its verified postcondition fails
  & Termination error (TE) & change the termination criterion \\
Source postcondition holds; a verified target precondition fails (readiness
error if the target also signaled \texttt{ready})
  & Handoff-condition violation (HCV) & add the missing precondition or
    alignment criterion \\
Both modules report success and all registered conditions pass; the composed
step still fails
  & Contract gap (CG) & register the omitted condition, unless the geometry
    check shows the entry region is execution-limited \\
\midrule
Entry handoff verified, preconditions hold throughout; failure inside the
skill's execution interval
  & Skill-internal execution (SIE) & improve the invoked skill; interface
    untouched \\
A condition that held at entry lapses mid-execution (user walks away,
occlusion appears)
  & Execution-time invariant lapse (EIL) & promote the condition to continuous
    monitoring \\
\bottomrule
\end{tabularx}
\end{table}

\paragraph{Two causes and a repairability criterion.}
Regrouped by repair action, the labels give the main result of this section.
H2 is the representative \emph{boundary-definition problem}:
\texttt{navigate\_to\_target} terminates on position alone while the greeting
requires facing the user, so the handoff breaks by construction---8 of 10
isolated attempts failed before the orientation criterion was
added\footnote{Counts from an isolated H2 test, not from the composed episodes
of \tabref{tab:fs-chain}.}---and the repair is to add the missing interface
convention. H5 is the representative \emph{executability problem}: every
condition is registered and passes, yet entry keeps failing, because geometry
binds against actuation.

\begin{figure}[!htbp]
\centering
\resizebox{\linewidth}{!}{\input{figures/handoff-two-causes.tex}}
\caption{The two causes at the end of navigation. At H2 the heading error is
never registered; at H5 the required start pose is declared but sits
$150^{\circ}$ away, in a band one step wide. Angles and offsets are
representative; band widths and the step are measured
(\tabref{tab:fs-rho}).}
\label{fig:fs-two-causes}
\end{figure}
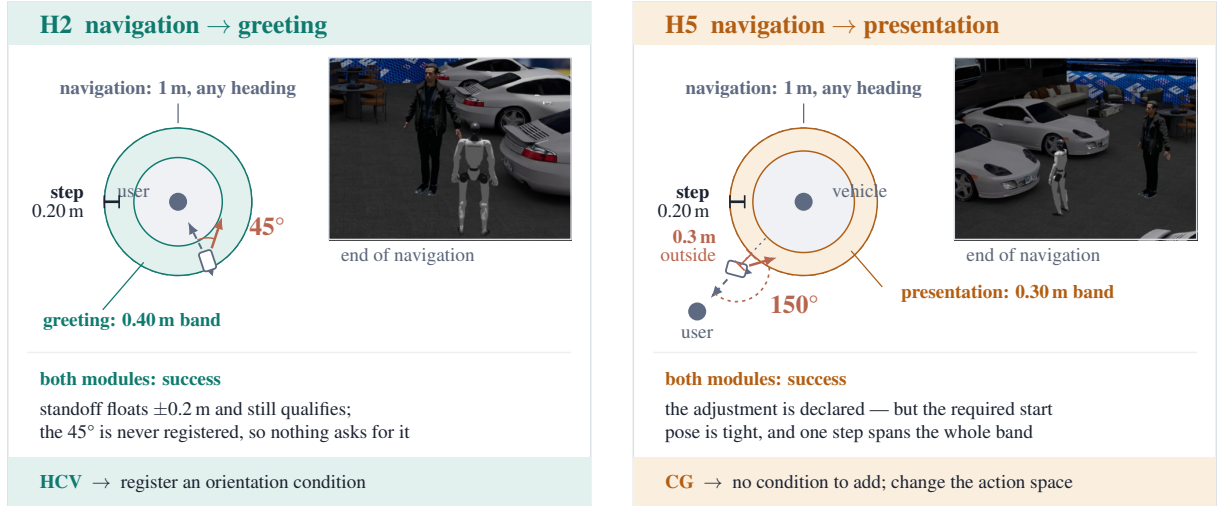

The criterion measures the feasible entry region in units of the minimum
executable step. If the region is narrower than one step, no correction can
stop inside it and the handoff is structurally unrepairable in the current
action space; if wider, correction is reliable only when the leftover slack
clearly exceeds execution error. Both quantities are measured---the step from
the actuator, the region from the downstream skill's rollout success
surface---so the criterion is decidable before execution. It also
disambiguates a contract gap, whose evidence pattern admits both causes: ample
slack means a condition was omitted, slack on the order of the execution error
means the entry region is execution-limited. Both handoffs share the same
$0.20$~m step: H2's $0.40$~m region leaves $0.20$~m of slack and the orientation
criterion repairs it, whereas field of view and occlusion squeeze H5's to
$0.30$~m and the remaining $0.10$~m is the same order as the execution error.
\tabref{tab:fs-rho} confirms this with a local adjustment budget (corrective
steps allowed after upstream termination to reach the downstream start pose):
at H2 a single step nearly saturates readiness, whereas H5 starts far lower and
needs the full five-step budget to catch up, each extra step re-rolling the same
error against the same slack. Figure~\ref{fig:fs-two-causes} draws the two
states at the end of navigation.

\begin{table}[!htbp]
\centering
\scriptsize
\setlength{\tabcolsep}{5pt}
\renewcommand{\arraystretch}{1.15}
\caption{Repairability criterion and readiness under a local adjustment budget.
The entry region (rollout success surface) and the minimum step (actuator) are
measured, and hence so is their ratio; Ready counts episodes whose handoff
state lands inside the calibrated entry region, not those whose registered
preconditions merely pass---at a contract gap the two differ.}
\label{tab:fs-rho}
\begin{tabular}{l cccc ccc}
\toprule
\multirow{2}{*}[-2pt]{Handoff}
& \multirow{2}{*}[-2pt]{Axis}
& \multirow{2}{*}[-2pt]{Region (m)}
& \multirow{2}{*}[-2pt]{Step (m)}
& \multirow{2}{*}[-2pt]{Steps wide}
& \multicolumn{3}{c}{Ready at budget} \\
\cmidrule(lr){6-8}
& & & & & 0 & 1 & 5 \\
\midrule
H2 navigation $\to$ greeting & approach & 0.40 & 0.20 & 2.0 & 12/20 & 16/20 & 17/20 \\
H5 navigation $\to$ presentation & approach & 0.30 & 0.20 & 1.5 & 6/15 & 11/15 & 13/15 \\
\bottomrule
\end{tabular}
\end{table}

\paragraph{Simulation study.}
We execute 20 composed episodes under varied start poses, approach directions,
referent ambiguity, occlusion, and user-confirmation behavior, at an adjustment
budget of one step (the budget-1 column of \tabref{tab:fs-rho}), and compare
automated attribution against human reference labels informed by privileged
simulator state. \tabref{tab:fs-chain} resolves the chain handoff by handoff:
the failure mass and the hidden subset---failures in which both adjacent
modules self-report success---concentrate on the two alignment handoffs.

\begin{table}[!htbp]
\centering
\scriptsize
\setlength{\tabcolsep}{5pt}
\renewcommand{\arraystretch}{1.15}
\caption{Per-handoff breakdown of the composed episodes: episodes arriving at
each handoff (Reached), breaking there (Failed), and failures with both
adjacent modules self-reporting success (Both~OK). Label abbreviations follow
\tabref{tab:fs-labels}; Abst is abstention; dashes denote zero.}
\label{tab:fs-chain}
\begin{tabular}{l ccc cccccc}
\toprule
\multirow{2}{*}[-2pt]{Handoff}
& \multirow{2}{*}[-2pt]{Reached}
& \multirow{2}{*}[-2pt]{Failed}
& \multirow{2}{*}[-2pt]{Both OK}
& \multicolumn{6}{c}{Attributed label} \\
\cmidrule(lr){5-10}
& & & & TE & HCV & CG & SIE & EIL & Abst \\
\midrule
H1 retrieval $\to$ nav.\ (person)   & 20 & --- & --- & --- & --- & --- & --- & --- & --- \\
H2 nav.\ $\to$ greeting             & 20 & 4 & 2 & 1 & 3 & --- & --- & --- & --- \\
H3 greeting $\to$ request           & 16 & 1 & 0 & --- & --- & --- & --- & 1 & --- \\
H4 resolution $\to$ nav.\ (viewpoint) & 15 & --- & --- & --- & --- & --- & --- & --- & --- \\
H5 nav.\ $\to$ presentation         & 15 & 4 & 3 & --- & --- & 3 & 1 & --- & --- \\
H6 presentation $\to$ termination   & 11 & 1 & 1 & --- & --- & --- & --- & --- & 1 \\
\midrule
Total (20 episodes, 10 completed)   & --- & 10 & 6 & 1 & 3 & 3 & 1 & 1 & 1 \\
\bottomrule
\end{tabular}
\end{table}

A negative control confirms discrimination: one H5 failure passes the entry
check like the contract gaps around it, but the trace shows the break occurring
mid-execution on the safety axis rather than pose, and attribution assigns it
to skill-internal execution, not a boundary.

\subsection{Real-Robot Evidence}
\label{sec:full-system-real}

We run the same task on the physical robot, with the
diagnostic restricted to hardware-observable inputs---module and tool status,
estimated robot state, controller events, onboard video, and odometry
pose---and human reference labels from the aligned trace, onboard video, and
on-site observation. Across 8 episodes, automated labels agree with the
reference on task outcome in every episode and on handoff attribution in 2 of
the 3 failed-or-partial ones, and every failure cause maps onto a class already
identified in simulation. Separately, in one episode the user interrupted
mid-skill without causing failure: control transfers outside all declared
handoffs, for which the interface defines no preemption semantics, and
responding would require duplex interaction---left to future work.

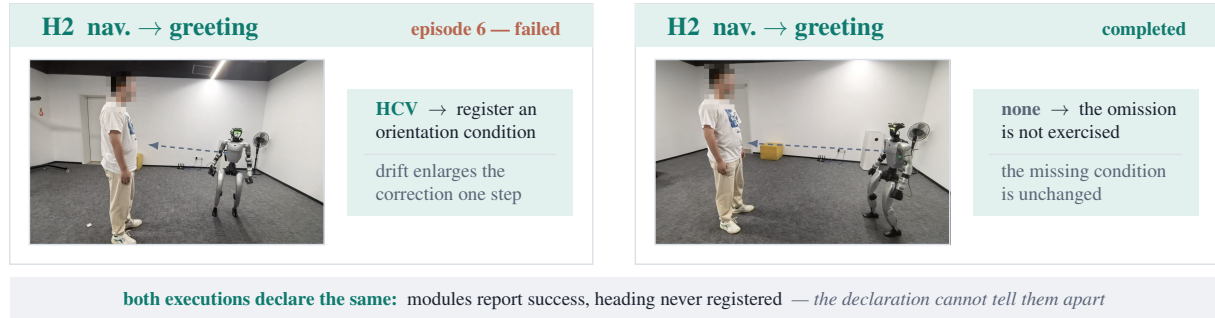
\begin{figure}[!htbp]
\centering
\resizebox{\linewidth}{!}{\input{figures/handoff-real-frames.tex}}
\caption{H2 on hardware at the end of navigation. The dashed line marks the
heading the greeting requires. Frames are representative; the entry region and
step of \tabref{tab:fs-rho} are calibrated in simulation, not re-measured on
hardware.}
\label{fig:fs-real-frames}
\end{figure}

One case shows the diagnostic resolving what module status alone cannot:
episode~6 fails at H2 with both modules self-reporting success, and attribution
returns a handoff-condition violation, citing [$t{=}39.3$\,s, $t{=}49.1$\,s].
Odometry drift, which hardware permits us to bound only to $0.10$--$0.20$~m,
enlarges the correction the robot must actually apply, leaving at most
$0.10$~m of the $0.20$~m slack that the one-step adjustment relies on in
simulation---no more than the margin at which H5 already fails.
\figref{fig:fs-real-frames} shows the state this resolves: the declared
evidence is the same whether the handoff breaks or completes.

\FloatBarrier

%% file: figures/handoff-two-causes.tex
%
%
\definecolor{HcBad}{HTML}{B8654F}%
\definecolor{HcTealFill}{HTML}{E3F1EE}%
\definecolor{HcOrangeFill}{HTML}{FAEEDF}%
\definecolor{HcTol}{HTML}{EFF1F5}%
\definecolor{HcRule}{HTML}{DCE0E7}%
\definecolor{HcRuleSoft}{HTML}{EDEFF3}%
\begin{tikzpicture}[
  x=0.18mm, y=-0.18mm,
  hcpanel/.style={draw=HcRule, line width=0.16mm},
  hcreq/.style={-{Latex[length=1.4mm, width=1.1mm]}, line width=0.22mm,
    draw=LocHandoff, dash pattern=on 0.72mm off 0.54mm},
  hcact/.style={-{Latex[length=1.4mm, width=1.1mm]}, line width=0.28mm,
    draw=HcBad},
  hctick/.style={draw=LocInk, line width=0.18mm},
]

\draw[hcpanel] (0,10) rectangle (356,318);
\fill[HcTealFill] (0,10) rectangle (356,36);
\node[anchor=base west, font=\scriptsize\bfseries, text=LocS1]
  at (12,29) {H2\enspace navigation $\to$ greeting};

\node[anchor=north west, inner sep=0] at (196,44)
  {\includegraphics[width=26.64mm]{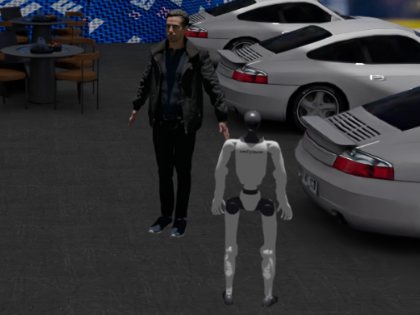}};
\draw[hcpanel] (196,44) rectangle (344,155);
\node[anchor=base west, font=\tiny, text=LocInkSoft] at (196,168)
  {end of navigation};

\node[anchor=base, font=\tiny\bfseries, text=LocHandoff] at (104,66)
  {navigation: 1\,m, any heading};
\draw[LocHandoff, line width=0.14mm] (104,72) -- (104,84);

\fill[HcTol] (104,132) circle[radius=45];
\draw[LocHandoff, line width=0.16mm, dash pattern=on 0.72mm off 0.54mm]
  (104,132) circle[radius=45];
\draw[HcTealFill, line width=3.24mm] (104,132) circle[radius=36];
\draw[LocS1, line width=0.16mm] (104,132) circle[radius=27];
\draw[LocS1, line width=0.16mm] (104,132) circle[radius=45];
\fill[LocHandoff] (104,132) circle[radius=5.5];
\node[anchor=base east, font=\tiny, text=LocInkSoft] at (94,129) {user};

\draw[hctick, line width=0.29mm] (59,132) -- (68,132);
\draw[hctick] (59,128) -- (59,136);
\draw[hctick] (68,128) -- (68,136);
\node[anchor=base east, font=\tiny\bfseries, text=LocInk] at (54,130) {step};
\node[anchor=base east, font=\tiny, text=LocInk] at (54,141) {0.20\,m};

\begin{scope}[rotate around={-71.6:(122,168)}]
  \draw[LocHandoff, line width=0.2mm, fill=white, rounded corners=0.3mm]
    (116,164) rectangle (128,172);
  \fill[LocHandoff] (128,166) -- (132,168) -- (128,170) -- cycle;
\end{scope}
\draw[hcreq] (118.4,160.8) -- (110,144);
\draw[hcact] (124.5,160.4) -- (131,141);
\draw[HcBad, line width=0.16mm]
  ([shift=(243.4:14)]122,168) arc[start angle=243.4, end angle=288.4, radius=14];
\node[anchor=base west, font=\scriptsize\bfseries, text=HcBad] at (140,152) {45\textdegree};

\draw[LocS1, line width=0.14mm] (58,198) -- (83,169);
\node[anchor=base west, font=\tiny\bfseries, text=LocS1] at (14,208)
  {greeting: 0.40\,m band};

\draw[HcRuleSoft, line width=0.16mm] (12,224) -- (344,224);
\node[anchor=base west, font=\tiny\bfseries, text=LocS1] at (12,244)
  {both modules: success};
\node[anchor=base west, font=\tiny, text=LocInk] at (12,262)
  {standoff floats $\pm$0.2\,m and still qualifies;};
\node[anchor=base west, font=\tiny, text=LocInk] at (12,276)
  {the 45\textdegree{} is never registered, so nothing asks for it};
\fill[HcTealFill] (0,288) rectangle (356,318);
\node[anchor=base west, font=\tiny, text=LocInk] at (12,307)
  {\textbf{\textcolor{LocS1}{HCV}}\enspace$\to$\enspace register an orientation condition};

\draw[hcpanel] (382,10) rectangle (738,318);
\fill[HcOrangeFill] (382,10) rectangle (738,36);
\node[anchor=base west, font=\scriptsize\bfseries, text=LocS0]
  at (394,29) {H5\enspace navigation $\to$ presentation};

\node[anchor=north west, inner sep=0] at (578,44)
  {\includegraphics[width=26.64mm]{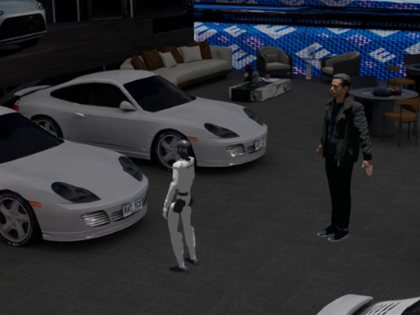}};
\draw[hcpanel] (578,44) rectangle (726,155);
\node[anchor=base west, font=\tiny, text=LocInkSoft] at (578,168)
  {end of navigation};

\node[anchor=base, font=\tiny\bfseries, text=LocHandoff] at (486,66)
  {navigation: 1\,m, any heading};
\draw[LocHandoff, line width=0.14mm] (486,72) -- (486,84);

\fill[HcTol] (486,132) circle[radius=45];
\draw[LocHandoff, line width=0.16mm, dash pattern=on 0.72mm off 0.54mm]
  (486,132) circle[radius=45];
\draw[HcOrangeFill, line width=2.52mm] (486,132) circle[radius=38];
\draw[LocS0, line width=0.16mm] (486,132) circle[radius=31];
\draw[LocS0, line width=0.16mm] (486,132) circle[radius=45];
\fill[LocHandoff] (486,132) circle[radius=5.5];
\node[anchor=base west, font=\tiny, text=LocInkSoft] at (496,129) {vehicle};

\draw[hctick, line width=0.29mm] (441,132) -- (450,132);
\draw[hctick] (441,128) -- (441,136);
\draw[hctick] (450,128) -- (450,136);
\node[anchor=base east, font=\tiny\bfseries, text=LocInk] at (436,130) {step};
\node[anchor=base east, font=\tiny, text=LocInk] at (436,141) {0.20\,m};

\begin{scope}[rotate around={-15:(445,173)}]
  \draw[LocHandoff, line width=0.2mm, fill=white, rounded corners=0.3mm]
    (439,169) rectangle (451,177);
  \fill[LocHandoff] (451,171) -- (455,173) -- (451,175) -- cycle;
\end{scope}
\draw[LocHandoff, line width=0.14mm, dash pattern=on 0.27mm off 0.45mm]
  (450.7,167) -- (464,153);
\draw[hcact] (452.7,171) -- (470,166);
\draw[hcreq] (439.4,179.4) -- (429,191);
\fill[LocHandoff] (421,199) circle[radius=5.5];
\node[anchor=base, font=\tiny, text=LocInkSoft] at (421,215) {user};
\draw[HcBad, line width=0.18mm, dash pattern=on 0.36mm off 0.45mm]
  ([shift=(-15.9:20)]445,173) arc[start angle=-15.9, end angle=133, radius=20];
\node[anchor=base west, font=\scriptsize\bfseries, text=HcBad] at (458,200) {150\textdegree};

\draw[HcBad, line width=0.18mm] (454,164) -- (447,171);
\draw[HcBad, line width=0.18mm] (451,161) -- (457,167);
\draw[HcBad, line width=0.18mm] (444,168) -- (450,174);
\node[anchor=base east, font=\tiny\bfseries, text=HcBad] at (440,157) {0.3\,m};
\node[anchor=base east, font=\tiny, text=HcBad] at (440,168) {outside};

\draw[LocS0, line width=0.14mm] (515,162) -- (534,184);
\node[anchor=base west, font=\tiny\bfseries, text=LocS0] at (538,191)
  {presentation: 0.30\,m band};

\draw[HcRuleSoft, line width=0.16mm] (394,224) -- (726,224);
\node[anchor=base west, font=\tiny\bfseries, text=LocS0] at (394,244)
  {both modules: success};
\node[anchor=base west, font=\tiny, text=LocInk] at (394,262)
  {the adjustment is declared --- but the required start};
\node[anchor=base west, font=\tiny, text=LocInk] at (394,276)
  {pose is tight, and one step spans the whole band};
\fill[HcOrangeFill] (382,288) rectangle (738,318);
\node[anchor=base west, font=\tiny, text=LocInk] at (394,307)
  {\textbf{\textcolor{LocS0}{CG}}\enspace$\to$\enspace no condition to add; change the action space};

\end{tikzpicture}%

%% file: figures/handoff-real-frames.tex
%
%
%
%
%
\definecolor{HcBad}{HTML}{B8654F}%
\definecolor{HcTealFill}{HTML}{E3F1EE}%
\definecolor{HcTol}{HTML}{EFF1F5}%
\definecolor{HcRule}{HTML}{DCE0E7}%
\begin{tikzpicture}[
  x=0.18mm, y=-0.18mm,
  hcpanel/.style={draw=HcRule, line width=0.16mm},
  hcsight/.style={-{Latex[length=1.2mm, width=0.9mm]}, line width=0.16mm,
    draw=LocHandoff, dash pattern=on 0.54mm off 0.43mm},
]

\draw[hcpanel] (0,10) rectangle (356,169);
\fill[HcTealFill] (0,10) rectangle (356,36);
\node[anchor=base west, font=\scriptsize\bfseries, text=LocS1]
  at (12,29) {H2\enspace nav.\ $\to$ greeting};
\node[anchor=base east, font=\tiny\bfseries, text=HcBad]
  at (344,29) {episode 6 --- failed};

\node[anchor=north west, inner sep=0] at (12,44)
  {\includegraphics[width=32.4mm]{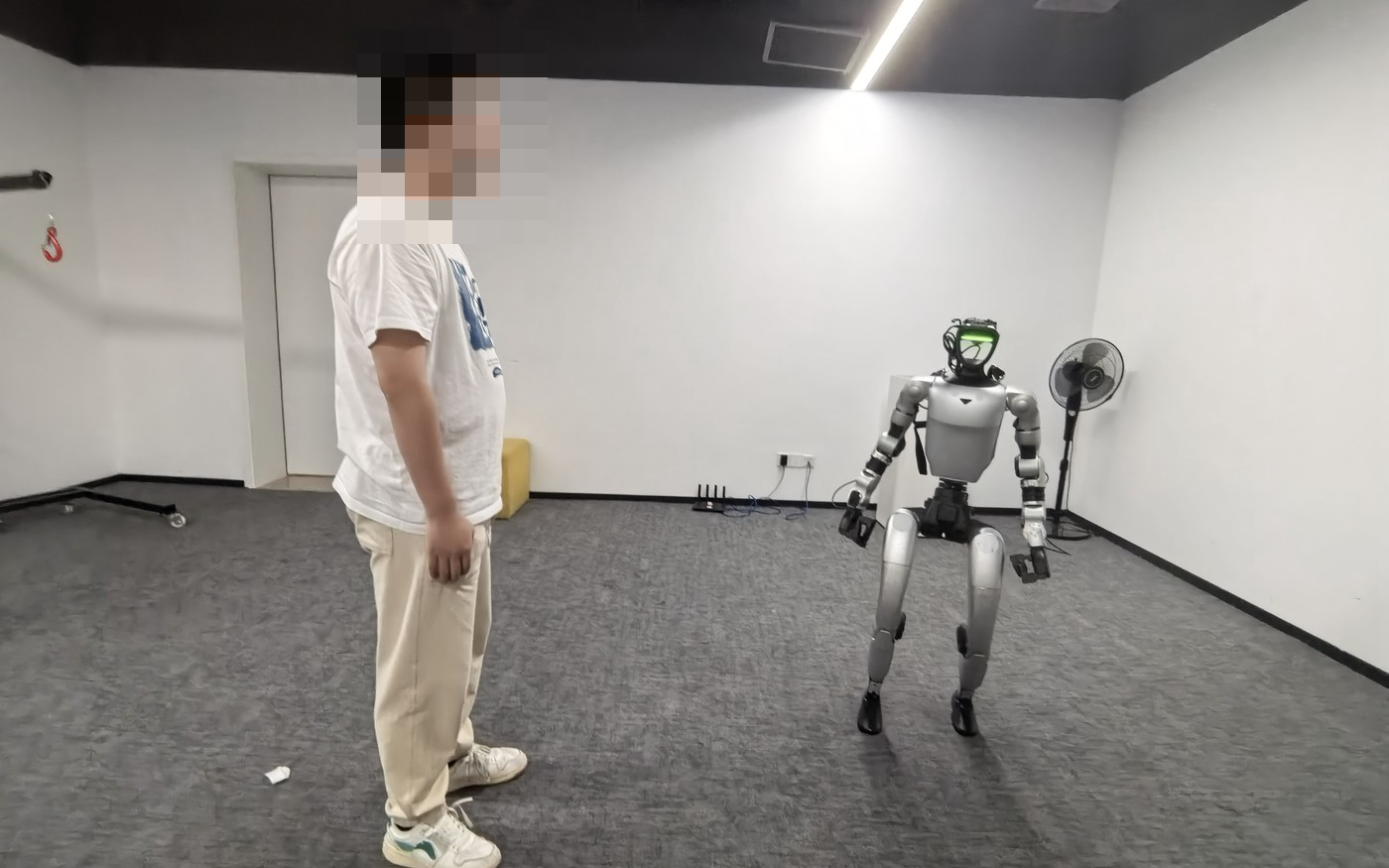}};
\draw[hcpanel] (12,44) rectangle (192,157);

\draw[hcsight] (133,101) -- (84,96);

\fill[HcTealFill] (206,62) rectangle (344,139);
\node[anchor=base west, font=\tiny, text=LocInk] at (216,78)
  {\textbf{\textcolor{LocS1}{HCV}}\enspace$\to$\enspace register an};
\node[anchor=base west, font=\tiny, text=LocInk] at (216,92)
  {orientation condition};
\draw[HcRule, line width=0.16mm] (216,102) -- (334,102);
\node[anchor=base west, font=\tiny, text=LocInkSoft] at (216,116)
  {drift enlarges the};
\node[anchor=base west, font=\tiny, text=LocInkSoft] at (216,130)
  {correction one step};

\draw[hcpanel] (382,10) rectangle (738,169);
\fill[HcTealFill] (382,10) rectangle (738,36);
\node[anchor=base west, font=\scriptsize\bfseries, text=LocS1]
  at (394,29) {H2\enspace nav.\ $\to$ greeting};
\node[anchor=base east, font=\tiny\bfseries, text=LocS1]
  at (726,29) {completed};

\node[anchor=north west, inner sep=0] at (394,44)
  {\includegraphics[width=32.4mm]{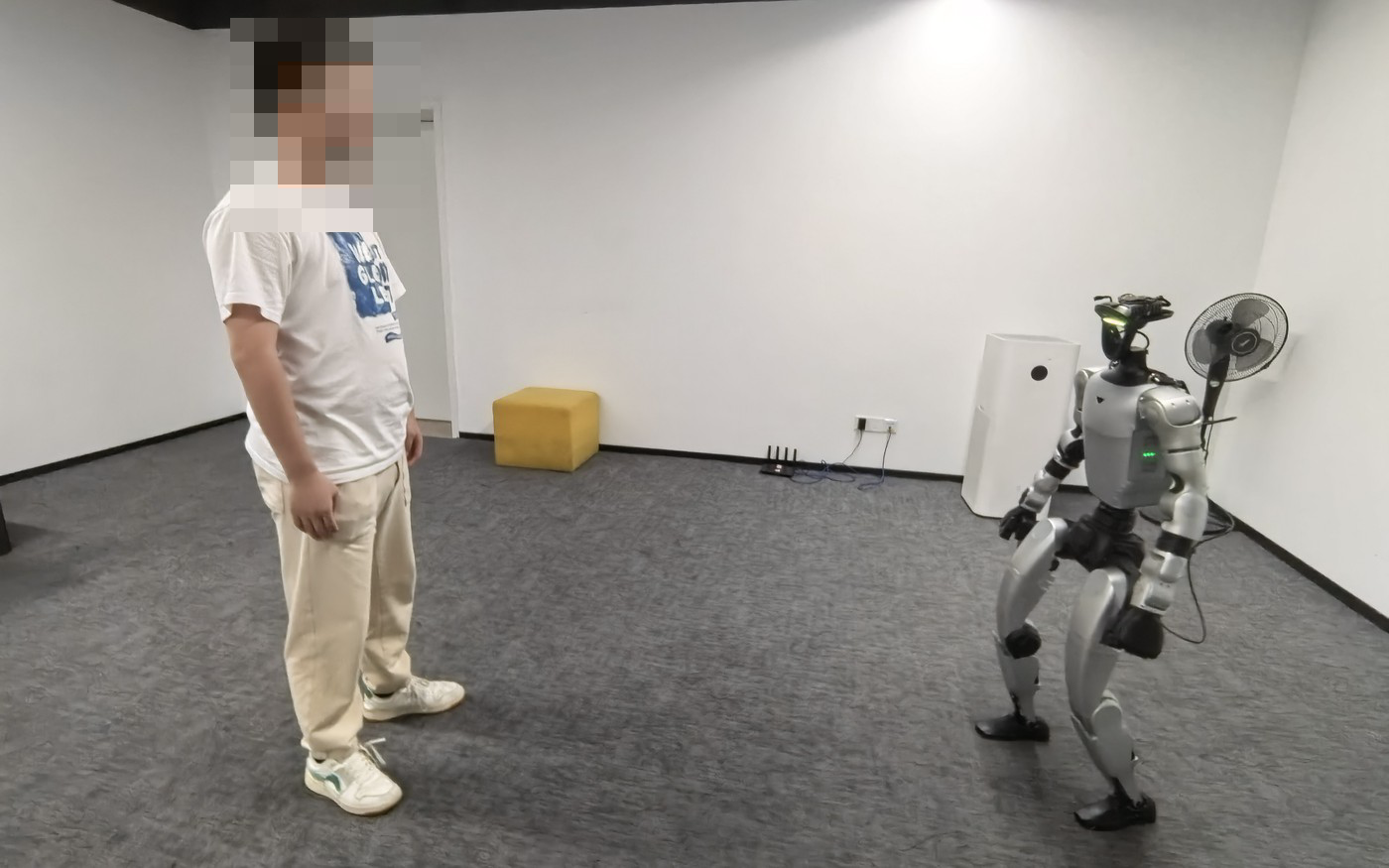}};
\draw[hcpanel] (394,44) rectangle (574,157);

\draw[hcsight] (536,100) -- (450,95);

\fill[HcTealFill] (588,62) rectangle (726,139);
\node[anchor=base west, font=\tiny, text=LocInk] at (598,78)
  {\textbf{\textcolor{LocInkSoft}{none}}\enspace$\to$\enspace the omission};
\node[anchor=base west, font=\tiny, text=LocInk] at (598,92)
  {is not exercised};
\draw[HcRule, line width=0.16mm] (598,102) -- (716,102);
\node[anchor=base west, font=\tiny, text=LocInkSoft] at (598,116)
  {the missing condition};
\node[anchor=base west, font=\tiny, text=LocInkSoft] at (598,130)
  {is unchanged};

\fill[HcTol] (0,177) rectangle (738,203);
\node[anchor=base, font=\tiny, text=LocInk] at (369,194)
  {\textbf{\textcolor{LocS1}{both executions declare the same:}}\enspace
   modules report success, heading never registered\enspace
   \textit{\textcolor{LocInkSoft}{--- the declaration cannot tell them apart}}};

\end{tikzpicture}%

%% file: main/conclusion.tex
\section{Conclusion}
\label{sec:conclusion}

We extended \di from a simulation-only substrate that quantified only the
foundation-model layer into one that quantifies the embodied layers---navigation
and manipulation (\sone) and whole-body control (\szero)---and carries a real
robot behind the same handle. The same three abstractions absorbed embodied
heterogeneity without a fourth. The distinctive return is that matched
simulated and physical rollouts share a parent trace identity while retaining
execution-domain-specific evidence. In System~0, this carries a standardized WBC
comparison from scalable simulation to matched physical evidence: the two
SONIC variants preserve their numerical ordering across seven matched
readings, while hardware increases tracking error and shifts joint dynamics
from lower RMS activity to larger upper-tail peaks.
At the full-system level, aligned trace evidence also turns
implicit skill-boundary assumptions into inspectable contracts. In the
simulated case study, five-way attribution maps failures to concrete repair
actions, and the measured ratio between a feasible entry region and the
minimum executable step separates interface-repairable boundary errors from
action-space limitations; on the physical robot, the same diagnosis
reproduces the failure classes identified in simulation and shows odometry
drift consuming the slack that the criterion budgets.